\documentclass[10pt,twocolumn,letterpaper]{article}

\usepackage{cvpr}      % To produce the REVIEW version
\usepackage{graphicx}
\usepackage{amsmath}
\usepackage{amssymb}
\usepackage{booktabs}
\usepackage{multirow}
\usepackage{bbding}
\usepackage[accsupp]{axessibility}

\usepackage{floatrow}
\usepackage[font=footnotesize,skip=0pt]{caption}
\usepackage[pagebackref,breaklinks,colorlinks]{hyperref}

\usepackage[capitalize]{cleveref}
\crefname{section}{Sec.}{Secs.}
\Crefname{section}{Section}{Sections}
\Crefname{table}{Table}{Tables}
\crefname{table}{Tab.}{Tabs.}

\def\cvprPaperID{2183} % *** Enter the CVPR Paper ID here
\def\confName{CVPR}
\def\confYear{2023}

\begin{document}

%%%%%%%%% TITLE - PLEASE UPDATE
\title{Dense Network Expansion for Class Incremental Learning}

\author{
Zhiyuan Hu$^1$~~~
Yunsheng Li$^2$~~~
Jiancheng Lyu$^3$~~~
Dashan Gao$^3$~~~
Nuno Vasconcelos$^1$~~~
\smallskip 
\\
$^1$UC San Diego~~~
$^2$Microsoft Cloud + AI~~~
$^3$Qualcomm AI Research~~~
\smallskip
\\
\{\tt\small{z8hu, nvasconcelos\}@ucsd.edu,~ yunshengli@microsoft.com,~ \{jianlyu, dgao\}@qti.qualcomm.com} 
}

%\author[1]{Alison Carefully}
%\author[2]{Ivor Question}
%\affil[1]{Department of Mathematics, University X}
%\affil[2]{Department of Biology, University Y}
%\affil[ ]{\textit {\{email1,email2,email3,email4\}@xyz.edu}}

%\author{Zhiyuan Hu\\
%UC San Diego\\
%{\tt\small z8hu@ucsd.edu}
% For a paper whose authors are all at the same institution,
% omit the following lines up until the closing ``}''.
% Additional authors and addresses can be added with ``\and'',
% just like the second author.
% To save space, use either the email address or home page, not both
%\and
%Second Author\\
%Institution2\\
%First line of institution2 address\\
%{\tt\small secondauthor@i2.org}
%}
\maketitle

%%%%%%%%% ABSTRACT
\begin{abstract}
    %The problem of {\it class incremental learning\/} (CIL) is considered. State-of-the-art approaches use a dynamic architecture based on network expansion (NE), in which a task expert sub-network is added per task. While effective from a computational standpoint, these methods lead to models that grow quickly with the number of tasks, which is unsustainable for many practical applications. A new NE  method, {\it dense network expansion\/} (DNE), is proposed to achieve a better trade-off between accuracy and model complexity. This is accomplished by the introduction of dense connections between the intermediate layers of the task expert networks, that enable the transfer of knowledge from old to new tasks via feature sharing and reusing. This sharing is implemented with a cross-task attention mechanism, based on a new {\it task attention block\/} (TAB), that implements information fusion across tasks. Unlike traditional attention mechanisms, TAB operates at the level of the feature mixing performed by the fully connected layers (MLP) of the transformer model, which it replaces. This is shown more effective than classical spatial attention for CIL.  The proposed DNE approach can strictly maintain the feature space of old classes while growing the network and feature scale at a much slower rate than previous methods. In result, it outperforms the previous SOTA methods by a margin of 4\% in terms of accuracy, with similar or even smaller model scale.
    The problem of {\it class incremental learning\/} (CIL) is considered. State-of-the-art approaches use a dynamic architecture based on network expansion (NE), in which a task expert is added per task. While effective from a computational standpoint, these methods lead to models that grow quickly with the number of tasks. A new NE  method, {\it dense network expansion\/} (DNE), is proposed to achieve a better trade-off between accuracy and model complexity. This is accomplished by the introduction of dense connections between the intermediate layers of the task expert networks, that enable the transfer of knowledge from old to new tasks via feature sharing and reusing. This sharing is implemented with a cross-task attention mechanism, based on a new {\it task attention block\/} (TAB), that fuses information across tasks. Unlike traditional attention mechanisms, TAB operates at the level of the feature mixing and is decoupled with spatial attentions. This is shown more effective than a joint spatial-and-task attention for CIL.  The proposed DNE approach can strictly maintain the feature space of old classes while growing the network and feature scale at a much slower rate than previous methods. In result, it outperforms the previous SOTA methods by a margin of 4\% in terms of accuracy, with similar or even smaller model scale.
\end{abstract}

%%%%%%%%% BODY TEXT
\vspace{-30pt}
\section{Introduction}
\label{sec:intro}
Deep learning has enabled substantial progress in computer vision. However, existing systems lack the human ability for continual learning, where tasks are learned incrementally. In this setting, tasks are introduced in sequential time steps $t$, and the dataset used to learn task $t$ is only available at the $t^{th}$ step. Standard gradient-based training is not effective for this problem since it is prone to \textit{catastrophic forgetting}: the model overfits on task $t$ and forgets the previous tasks. This is unlike humans, who easily learn new tasks without forgetting what they know. While continual learning can be posed for any topic in computer vision, most research has addressed classification and the {\it class incremental\/} (CIL) setting~\cite{lwf}. In CIL, tasks consist of subsets of disjoint classes that are introduced sequentially. Most approaches also allow the learning of task $t$ to access a small buffer memory of examples from previous tasks.  

%Different strategies have been proposed to solve the CIL problem. Distillation methods\cite{lwf, icarl, lucir, podnet, darker, afc, dytox, ebll, lvt} assume that, once trained to solve task $t$, the network should produce the same logits, or similar intermediate features, for classes from previous tasks. Parameter regularization methods\cite{ewc, si} limit the variation of model parameters. They estimate the importance of each parameter to the previous tasks and ensure that important parameters do not change substantially. Gradient methods~\cite{gem, nscl, adns} estimate the null space of the existing feature space and project the gradients of task $t$ into this null space, so that the feature vectors generated for examples of the previous tasks remain approximately the same. These methods try to fit all tasks into a single model, preserving properties of the feature space from one task to the next. This is, however, difficult to guarantee due to the scarcity of prior task data. Furthermore, as the number of tasks grows, the model will eventually run out of capacity to accommodate new tasks. 

Different strategies have been proposed to solve the CIL problem. \textit{Distillation methods}\cite{lwf, icarl, lucir, podnet, darker, afc, dytox, ebll, lvt} and \textit{parameter regularization} methods\cite{ewc, si} regulate the new model by the output logits, intermediate features or important parameters. Gradient methods~\cite{gem, nscl, adns} estimate the null space of the existing feature space and project the gradients of the next task into this null space, so that the newly learned features are orthogonal to the previous ones. These methods try to fit all tasks into a single model, preserving properties of the feature space from one task to the next. This is, however, difficult to guarantee due to the scarcity of prior task data. Furthermore, as the number of tasks grows, the model will eventually run out of capacity to accommodate new tasks. 

\begin{figure*}[t]\RawFloats
\centering
\begin{tabular}{cc}
    \includegraphics[width=.54\linewidth]{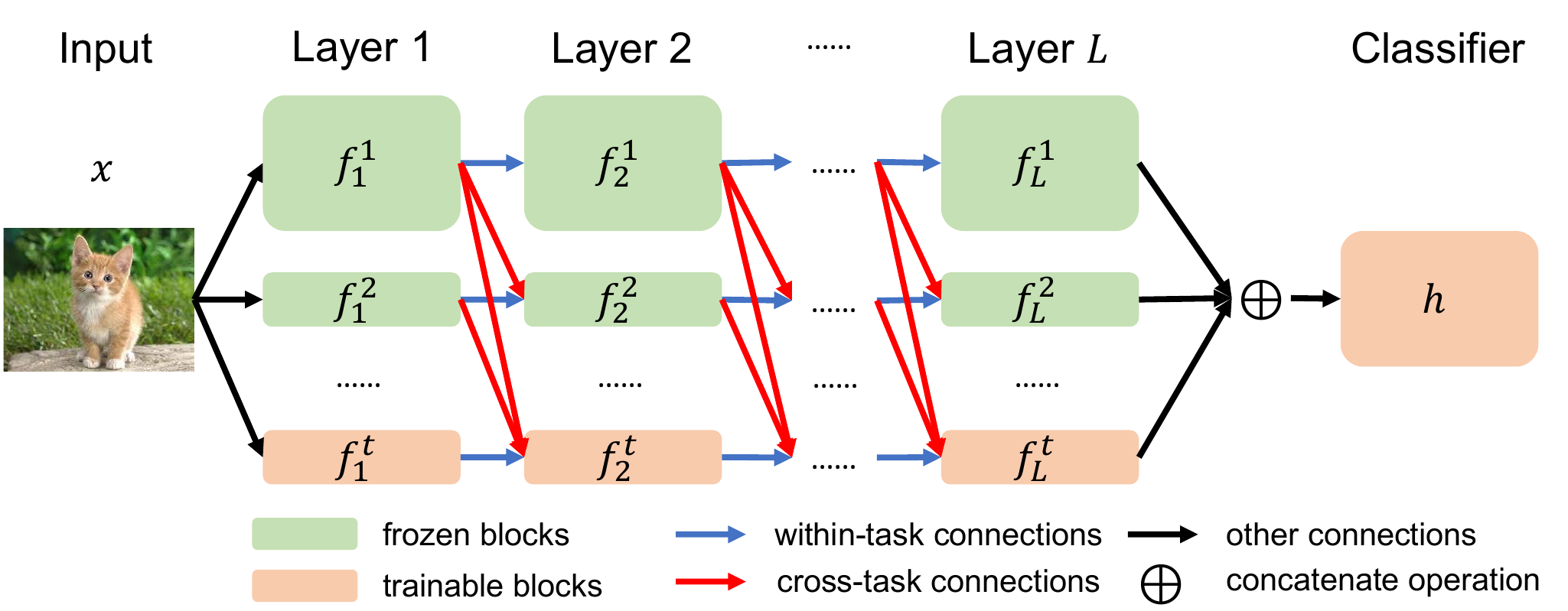} &  
    \includegraphics[width=.27\linewidth]{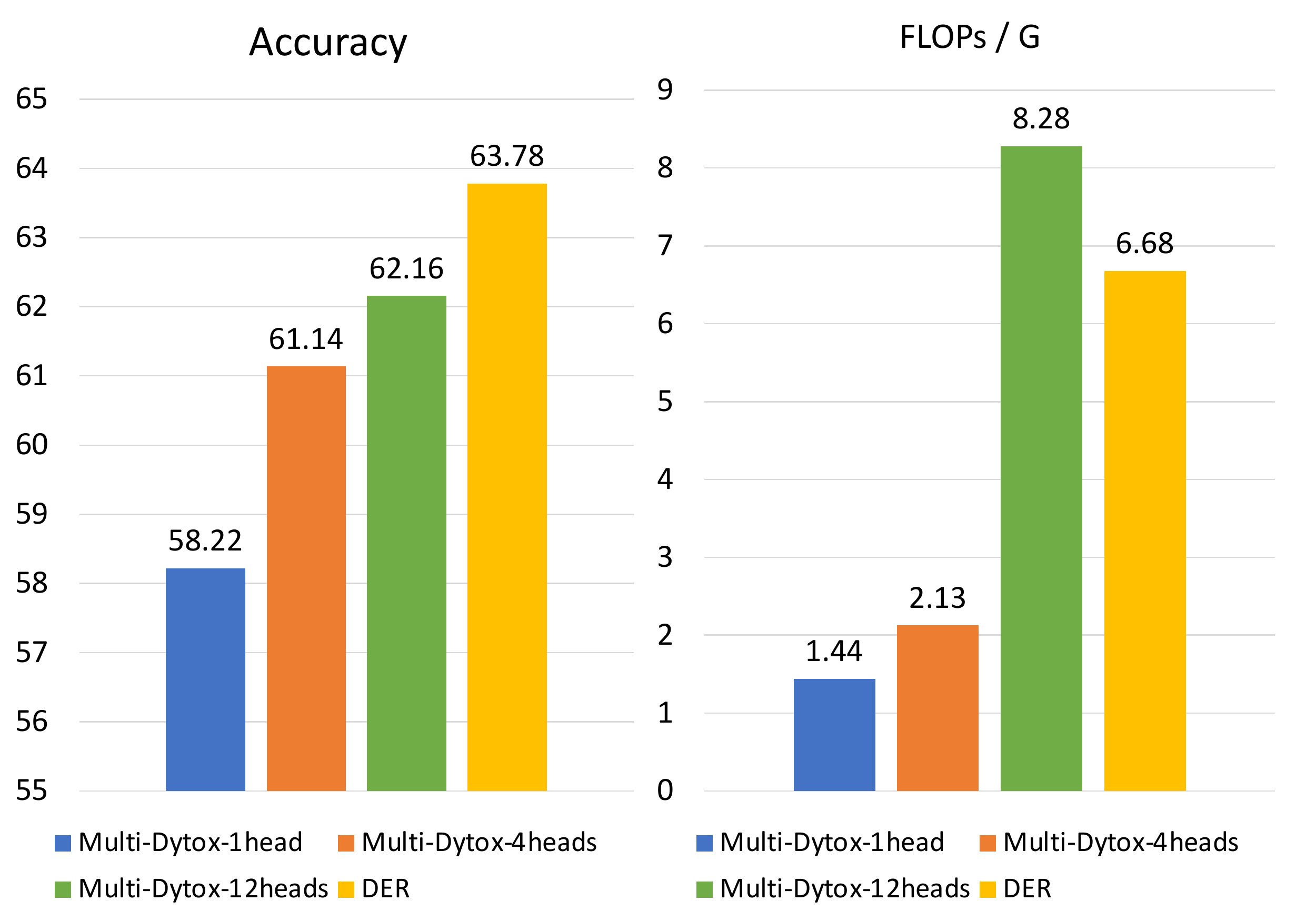}
\end{tabular}
\caption{CIL by NE. Left: Given a new task $t$, a new branch, denoted the {\it task $t$ expert}, is added while freezing existing experts. In classical NE, the model is simply replicated per task, originating a set of $t$ independent models, whose outputs are concatenated into a classifier. In the proposed DNE scheme, a cross-task attention mechanism (red connections) is introduced to allow the re-use of knowledge from previous experts and smaller experts per task. Right: Comparison of model accuracy and size for various implementations of NE, using DER~\cite{der} and multi-Dytox~\cite{dytox} models of different sizes per task expert, on the CIFAR100 dataset. Both accuracy and FLOPs are shown for the final model, which classifies $100$ classes grouped into $6$ tasks.}
\label{fig:model}
\end{figure*}

%Dynamic architecture methods~\cite{expert, rpsnet, der, pretrain, foster} address these problems by freezing the model that solves the previous tasks and adding a new subnetwork per task, i.e. by performing a  {\it network expansion} (NE) per task. As illustrated in Figure~\ref{fig:model}, a network $f^1$ learned for task $1$ is augmented with new networks $f^t$, denoted as {\it task experts,\/} for each subsequent task $t$, and the feature spaces concatenated. Since the original features are always accessible for examples from previous classes, these methods achieve the best performances on CIL benchmarks.  However, the process is very inefficient in terms of both model size and complexity. The right side of the figure shows the accuracy and size of several NE models on the CIFAR100 dataset. Best accuracies are obtained with larger networks and the model grows very quickly with the number of tasks. For most practical applications, this rate of growth is unsustainable. 

\textit{Network expansion} (NE) methods~\cite{expert, rpsnet, der, pretrain, foster} address these problems by freezing the model that solves the previous tasks and adding a new subnetwork per task. As illustrated in Figure~\ref{fig:model}, a network $f^1$ learned for task $1$ is augmented with new networks $f^t$, denoted as {\it task experts,\/} for each subsequent task $t$, and the feature spaces concatenated. \textit{NE with cross connections} (NEwC) methods~\cite{pnn, clnp, packnet} further add links across tasks (red lines in Figure~\ref{fig:model}) to further transfer knowledge from old to new tasks. Since the original features are always accessible for examples from previous classes, these methods achieve the best performances on CIL benchmarks. However, the process is very inefficient in terms of both model size and complexity. The right side of the figure shows the accuracy and size of several NE models on the CIFAR100 dataset. Best accuracies are obtained with larger networks and the model grows very quickly with the number of tasks. For most practical applications, this rate of growth is unsustainable.

While NE and NEwC achieve state of the art performance among CNN-based CIL methods, we show that they do not translate well to the more recent transformer architecture~\cite{vit}. Standard transformers learn spatial connections across image patches through a spatial attention mechanism. A natural CIL extension is to feed the input image to multiple heads, each corresponding to a task. Attention can then be computed between all image patches of all heads,  leading to a Spatial-and-Task Attention (STA) mechanism. This strategy has is commonly used in multi-modal transformers~\cite{clip, frozen}. However, in CIL, the features generated per patch by different heads are extracted from exactly the same image region. Hence, their representations are highly similar and STA is dominated by the attention between replicas of the same patch. Furthermore, because all patches are processed by all heads, the remaining attention is dispersed by a very large number of patch pairs. This leads to the {\it fragmentation\/} of attention into a large number of small-valued entries, which severely degrades performances. To overcome 
 this problem, we propose a \textit{Dense Network Expansion} (DNE) strategy that disentangles spatial and cross-task attention.  Spatial attention is implemented by the standard transformer attention mechanism. Cross-task attention is implemented by a novel {\it task attention block\/} (TAB), which performs attention at the level of feature-mixing, by replacing the multi-layer perceptron (MLP) block with an attention module.

Overall, the paper makes four contributions. First, we point out that existing NE methods are unsustainable for most practical applications and reformulate the NE problem, to consider the trade-off between accuracy and model size. Second, we propose the DNE approach to address this trade-off, leading to a CIL solution that is both accurate and parameter efficient. Third, we introduce an implementation of DNE based on individual spatial and cross-task attentions. Finally, extensive experiments show that DNE outperforms all previous CIL methods not only in terms of accuracy, but also of the trade-off between accuracy and scale.   

%\begin{enumerate}
%    \item We propose the Dense Network Expansion method to solve class incremental learning in a both effective and efficient manner.
%    \item We design the task attention block to transfer knowledge from old tasks to the new task.
%    \item We conduct extensive experiments to show the effective and efficient of the proposed DNE method.
%\end{enumerate}

\vspace{-5pt}
\section{Related Works}
\label{sec:relate}

CIL aims to learn a sequence of classification tasks, composed of non-overlapping classes, without catastrophic forgetting. 
The literature can be roughly divided into \emph{single model} and \emph{network expansion} (NE) methods. 

\noindent{\bf Single model methods:}
Single model methods assume that catastrophic forgetting can be avoided by imposing constraints on the model as new tasks are learned.  \emph{Distillation} methods\cite{lwf, icarl, lucir, dytox, podnet, afc, ebll, lvt} feed the data from the current task to both old and new networks. \cite{lwf, lucir, icarl, dytox} force the new network to reproduce the  logits of the old network, to guarantee stable logits for old class data. \cite{podnet} and \cite{afc} further match the intermediate feature tensors of the old and new networks. \cite{lvt} proposes a transformer-based method that defines an external key feature per attention block to represent old tasks and regularizes this feature. \emph{Parameter regularization} methods\cite{ewc, si} assume that the knowledge of old classes is stored in the model parameters. For example, \cite{ewc} uses the Fisher information matrix of a parameter to estimate its importance for certain classes. Important parameters are then kept stable across tasks using a L2 regularization weighted by parameter importance. \emph{Gradient} methods\cite{gem, nscl, adns} modify gradient descent training to eliminate catastrophic forgetting. \cite{gem} computes loss gradients for current and old classes. The current gradient is then forced to have a positive dot product with the old gradients, so that training of current classes does not increase the loss for old ones. Since this is typically difficult to satisfy, \cite{nscl} proposes to directly estimate the null space of the gradients of old classes. The new gradients are then projected into this null space, to eliminate their influence on old classes.

\noindent{\bf NE methods:} While single model methods solve the catastrophic forgetting problem to some extent, the scarcity of data from previous tasks makes it difficult to guarantee model stability across tasks. Furthermore, a single model eventually lacks the capacity to accommodate all tasks, as task cardinality grows. \emph{Dynamic architecture} methods\cite{expert, rpsnet, der, pretrain, foster} address this problem with NE. A new network, or task expert, is learned per task, while previous experts are frozen.  In a result, the features originally produced for old task classes are always available. \cite{expert,der} train an entire backbone per task expert. \cite{expert} uses an auto-encoder task level selector, to select the expert that best suits the example to classify. \cite{der} directly concatenates the feature vectors generated by old and new experts for classification. These methods outperform single model ones, but lead to models that grow very quickly with the number of tasks. Hence, they are unrealistic for most practical applications. 

Some works have sought a better trade-off between model accuracy and complexity. \cite{pretrain} notes that the shallow layers of the old and new experts learn similar low-level features, and uses pre-trained shallow layers that are shared by all experts. This reduces model size and leverages the power of pretraining. \cite{foster} adds an entire backbone per task, but distills the old and new networks into a single one, with the size of a single backbone. The model size is thus kept fixed as the number of tasks grows. However, these solutions are mostly pre-defined, either breaking the model into a single model component (early layers)  and a component that grows without constraints (later layers) or effectively using a single model. The proposed DNE approach instead explores the use of connections between task experts, feature reuse between tasks, and dynamic information fusion through cross-task attention to enable a significantly better trade-off between accuracy and model size.

\vspace{-10pt}
\section{Dense Network Expansion}

In this section, we first revisit the mathematical definition of CIL problem and previous NE methods amd then introduce the DNE approach.
\vspace{-10pt}
\paragraph{\bf CIL:}
We consider the problem of image classification, where image $x$ has category label $y$, and the CIL setting, where a model $g(x;\theta)$ of parameter $\theta$ learns a sequence of $M$ tasks $T = \{T_1, T_2, \dots, T_M\}$. Each task $T_i$ has a dataset $D_i=\{(x_k, y_k)\}_{k=1}^{N_i}$ of $N_i$ samples from a class set $\mathcal{Y}_i$ disjoint from the remaining task sets, i.e. $\mathcal{Y}_i \cap \mathcal{Y}_j=\varnothing, i\neq j$. 

At step $t$, the model predicts the posterior class probabilities $g_i(x;\theta)=P_{Y|X}(i|x)$ for all the classes observed so far $i\in\mathcal{Y}_1\cup\mathcal{Y}_2\cup\dots\cup\mathcal{Y}_t$. In strict incremental learning, only the dataset $D_t$ of the current task is available for learning. However, this has proven too challenging, most incremental learning methods assume the availability of a small memory buffer $\mathcal{M}=\{(x_k, y_k)\}_{k=1}^S$ with data from prior tasks, where $S$ is the buffer size and each sample $(x_k, y_k)$ is drawn from the union of previous datasets $D_1\cup D_2\cup\dots\cup D_{t-1}$.

Most deep learning models $g(x;\theta)$ consist of a series of blocks and a classifier. The network parameters can be divided into $\theta=\{\theta_1, \theta_2, \dots, \theta_L, \phi\}$ where $\theta_i$ is the parameter vector of the $i$-th block, $\phi$ that of the classifier and $L$ the number of blocks. The network is then implemented as
\begin{eqnarray}
    r_0& =& x\\ 
    r_l &=& f_l(r_{l-1}; \theta_l), \,\,\, l \in \{1, \ldots, L\} \label{eq:fl} \\
    g(x;\theta) &=& h(r_L; \phi),
\end{eqnarray}
where $f_l$ represents the $l$-th block, $h$ is the classifier and $r_l$ the feature tensor at the output of the $l$-th block.
\vspace{-8pt}
\paragraph{\bf NE methods:}
A popular approach to CIL is to rely on the NE procedure of Figure~\ref{fig:model}, which learns a task expert $f^t$, of the form of (\ref{eq:fl}), per task $t$. To learn this branch, NE methods freeze experts $f^1, \ldots, f^{t-1}$ of the previous tasks, forcing the new expert to generate features specific to task $t$.  The entire network can then be written as 
\begin{eqnarray}
    r_0^t &=& x\\ 
    r_l^t &=& f^t_l(r_{l-1}^t; \theta_l^t), , \,\,\, l \in \{1, \ldots, L\} \label{eq:rl} \\
    g(x;\theta) &=& h(r_L; \phi), \label{eq:g}
\end{eqnarray}
where $l$ denotes block, $t$ denotes task, 
\begin{eqnarray}
    r_l &=& r_l^1\oplus r_l^2\oplus\dots\oplus r_l^t\\
    \theta_l &=& \{\theta_l^1, \theta_l^2, \dots, \theta_l^t\} \\
    \phi &=& \{\phi^1, \phi^2, \dots, \phi^t\}
\end{eqnarray}
and $\oplus$ is the concatenation operation. The parameters $\theta_l^1, \dots,\theta_l^{t-1}$ are frozen and only $\theta_l^t$ is learned for each block $l$. The classifier parameters $\phi$ are learned over the entire feature vector $r_L$, leveraging the features produced by all task experts to assign $x$ to one of the classes from all tasks.

%parameters $\theta$ and then assign a new set of parameters $\hat{\theta}_l$ for each block. The output of each block is also expanded by an extra part $\hat{r}_l$. Since the old output $r_i$ only depends on $\theta_l$ and $r_{l-1}$, they will also be fixed and is identical for the old and new network. We can thus divide the block parameters $\theta_i$ and output $r_l$ into: where $\theta_l^j$ and $r_l^j$ are the parameters and output learned in task $j$, $t$ is the task number and .

\vspace{-8pt}
\paragraph{\bf Challenges:}
While NE is popular, it can be quite inefficient. If the dimensionality of the new task parameters $\theta_l$ is small, i.e. the new task expert is a small network, the model has limited capacity to learn the new task and the recognition accuracy can be low. On the other hand, if the dimensionality is large, e.g. a full network as in the popular DER~\cite{der} method, the model size grows very quickly and there can be too much capacity, leading to overfitting. 

This problem is illustrated in the right side of Figure~\ref{fig:model}, which shows both the accuracy and model size obtained by expanding the Dytox model of~\cite{dytox} using various parameter dimensionalities. In this example, each expert is a Dytox transformer with a different number of spatial attention heads $k$. The figure shows the CIL performance on CIFAR100, for a sequence of 6 tasks of $\{50, 10, 10, 10 ,10, 10\}$ classes each,  by networks with $k \in \{1, 4, 12\}$. While a single attention head per task is not sufficient to enable high recognition accuracy, the size of the model grows very quickly as more heads are used per task, without a large increase in recognition accuracy. 
%For reference, we also show the accuracy and size of the CNN-based DER model. In this example, the transformer-based architecture is never able to outperform the latter.
In this work, we seek expansion methods with a better trade-off between model accuracy and size.

\begin{figure}[t]\RawFloats
    \centering
    \includegraphics[width=0.9\linewidth]{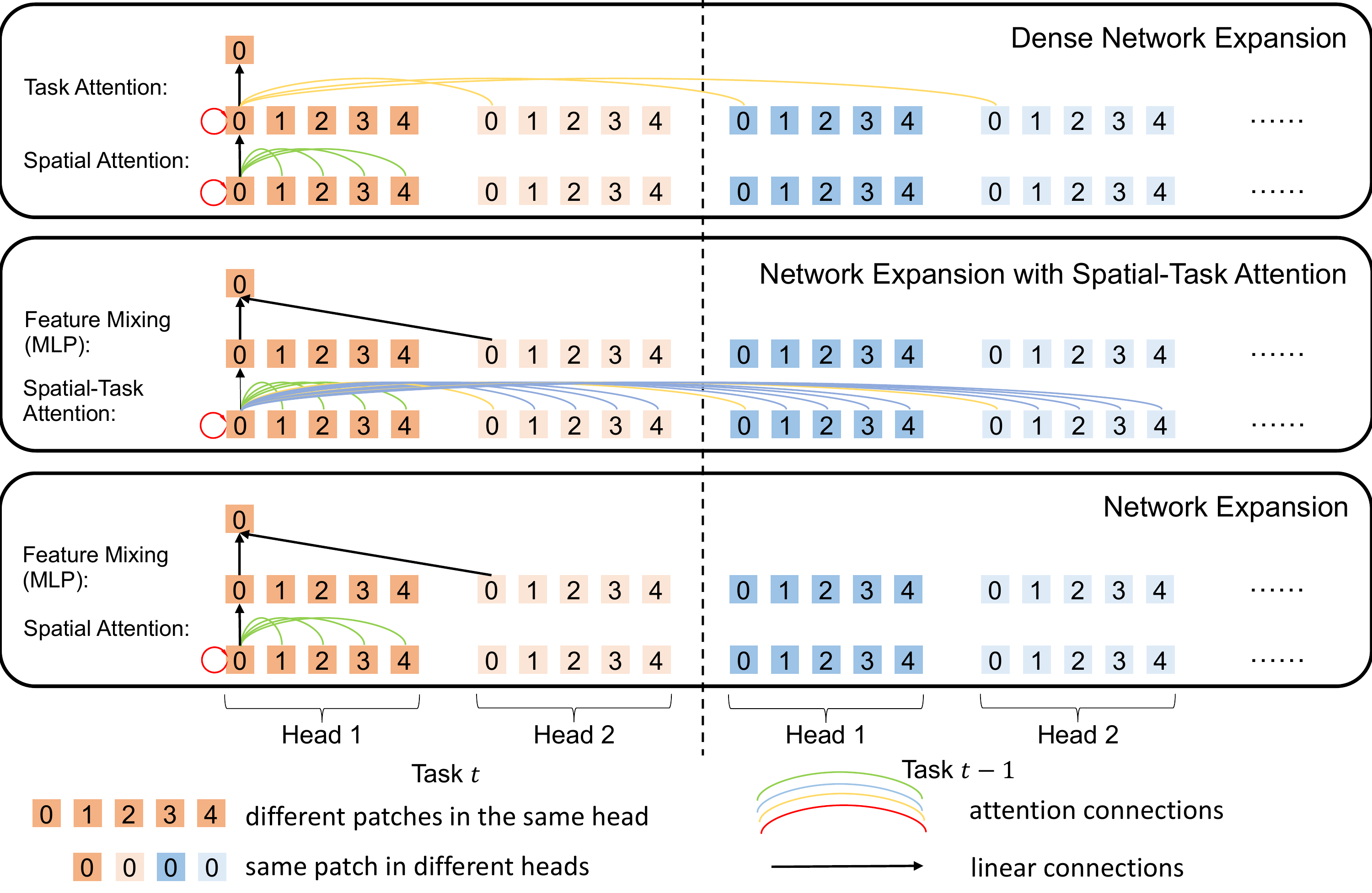}
    \caption{Different implementations of cross-task attention. Only one output patch is shown for simplicity. Bottom: network expansion has no cross-task attention. Middle: Network expansion with Spatial Attention adds spatial attention connections between task experts. Top: DNE replaces the MLP with cross-task attention connections between corresponding patches of different heads.}
    \label{fig:teaser}
\end{figure}

We hypothesize that the inefficiency of NE is due to its lack of ability to transfer knowledge from old tasks to new task. To address this problem we introduce {\it cross-task connections,\/} connecting the layers of the new task expert to those of the experts already learned for the previous tasks. For task $t$, block $f_l^t$ takes feature vectors of all tasks as input to generate the outputs of the task,
\begin{equation}
    r_l^t = f_l^t(r_{l-1}^1, \dots, r_{l-1}^{t-1}, r_{l-1}^t; \theta_l^t)
    \label{eq:rlt}
\end{equation}
The question becomes how to design a task $t$ expert capable of learning features {\it complimentary\/} to those already available. For this, the expert must integrate the features generated by the previous task experts. 
This issue has been considered in the literature for purely convolutional models \cite{clnp, pnn, packnet}, where cross-task connections are trivially implemented as a convolution block that linearly projects the output channels of all task experts into input channels of the next block of task $t$, %{\color{red} SHOULDNT THE SUBSCRIPT BE l-1 INSIDE THE CONVOLUTION?} {\color{blue} FIXED}
\begin{equation}
    r_l^t = \text{Conv}(r_{l-1}^1\oplus r_{l-1}^2\oplus\dots\oplus r_{l-1}^t)
    \label{eq:cross_conv}
\end{equation}
However, this architecture underperforms NE approaches such as DER~\cite{der}. In this work, we revisit the question in the context of transformer models, which have stronger ability to transfer information across the features of a network layer, via attention mechanisms. This has been used to integrate information across data streams produced by either different image regions~\cite{vit, detr, maskformer} or different perceptual modalities~\cite{clip}. However, as we will next see, cross-task connections are not trivial to implement for this model.

\vspace{-10pt}
\paragraph{\bf Independent Attention Model:}
We leverage the Vision Transformer (ViT)~\cite{vit} as backbone used to implement all task expert branches. Under the ViT architecture~\cite{transformer}, block $f_l$ of layer $l$ implements a sequence of two operations
\begin{align}
    s_l &= r_{l-1} + \text{MHSA}_l(\text{LN}(r_{l-1})) \label{eq:sl_transf} \\
    r_l &= s_l + \text{MLP}_l(\text{LN}(s_l)) \label{eq:rl_transf}
\end{align}
where MHSA is a multi-head self-attention block, MLP a multi-layer perceptron, and LN a layer normalization. Under the NE strategy, a transformer  is added independently per task. Since, as illustrated at the bottom of Figure~\ref{fig:teaser}, this has no connections between different task branches, we denote it as the {\it independent attention\/} (IA) model. As shown in the right part of Figure~\ref{fig:model}, IA does not achieve high CIL accuracy unless each task expert is a computationally heavy multi-headed transformer. We seek to introduce the inter-task connections of (\ref{eq:rlt}) to improve the performance of the simplest models, namely that with a single head per task.
%, but only for very large models.

\begin{figure}\RawFloats
    \centering
    \includegraphics[width=0.9\linewidth]{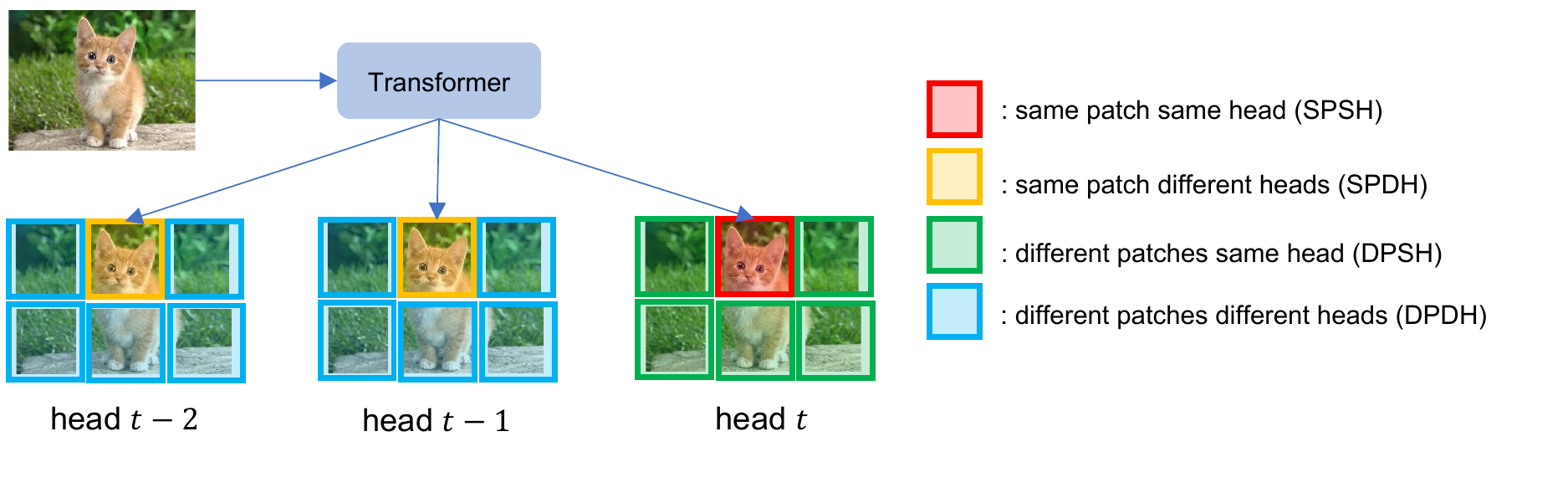}
    \caption{Four different types of tokens in STA when computing the output of top-middle patch in head $t$ (token in red frame)}
    \label{fig:fourtype}
\end{figure}

\vspace{-8pt}
\paragraph{\bf Spatial Task-Attention Model:}
The straightforward generalization of (\ref{eq:cross_conv}) to the transformer model is to implement a joint attention block across tokens of different tasks. This is illustrated in the middle of Figure~\ref{fig:teaser} and denoted as {\it Spatial-Task Attention\/} (STA). Omitting the block index $l$ for simplicity, (\ref{eq:sl_transf}) is replaced by 
\begin{align*}
    s_{p}^t = r_p^t + \text{SA}(\text{LN}(&\{r_{p}^0\}_{p=0}^{P-1}, \{r_{p}^1\}_{p=0}^{P-1},\dots, \{r_{p}^{t-1}\}_{p=0}^{P-1}))
\end{align*}
where $p$ is a patch index, $P$ the total number of patches, and $t$ the task branch. The MLP of (\ref{eq:rl_transf}) is maintained.

Our experiments show that this approach is not effective. Unlike most other uses of transformers, say multimodal models like CLIP~\cite{clip}, the different branches of the CIL network (task experts) process the {\it same input\/} image. As illustrated in Figure~\ref{fig:fourtype}, the image to classify is fed to {\it all\/} experts. Hence, the cross-task attention connects different projections of the {\it same\/} patches. Without lack of generality, the figure assumes each task expert is implemented with a single transformer head.  Consider the processing of the patch in red by expert $t$ and note that the yellow patches processed by the prior task experts all refer to the same image region. In result, the dot-products between the projections of red and yellow patches dominate the attention matrix. 

\begin{figure}\RawFloats
    \centering
    \includegraphics[width=\linewidth]{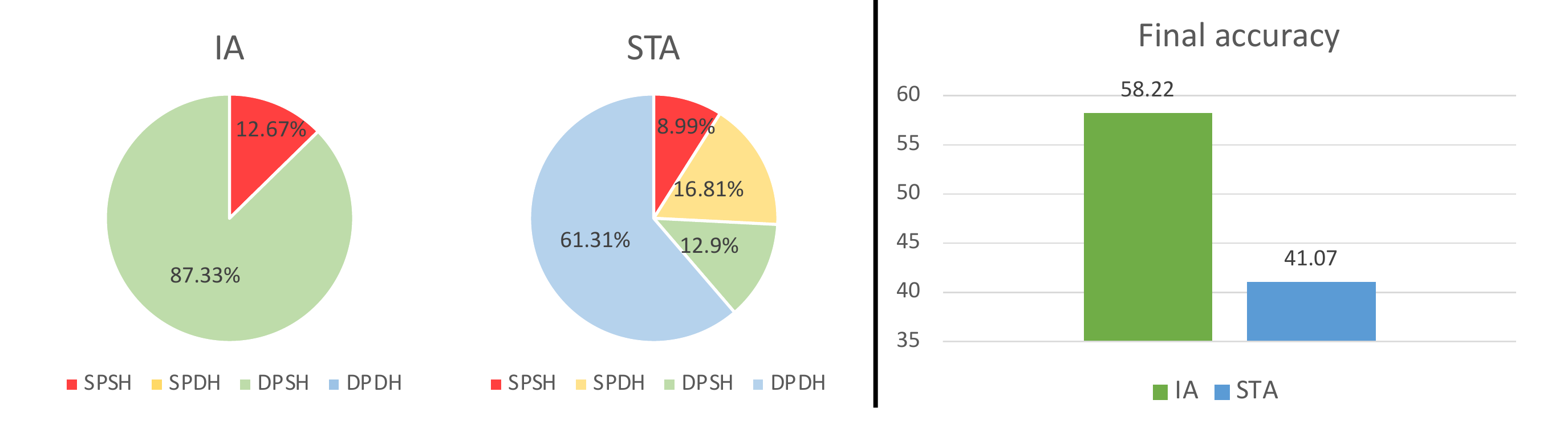}
    \caption{Left: Distribution of attention by the four groups of Figure~\ref{fig:fourtype}, for IA and STA models. Right: Final accuracy of IA and STA.}
    \label{fig:portion}
\end{figure}

As defined in the figure, there are four types of attention dot-products: SPSH (red patch with itself), SPDH (red-yellow), DPSH (red-green) and DPDH (red-blue) . The left of figure~\ref{fig:portion} compares the percentage of attention of each of these types of the STA model to those of the IA model, which only has SPSH and DPSH connections. Connections to the same patch (either red-red or red-yellow) account for 25.8\% of the STA attention strength, as opposed to only 12.7\% for the IA model. Furthermore, because STA also includes the DPDH connections to the projections of all blue patches of Figure~\ref{fig:fourtype} (rather than just the DPSH of those in green), the spatial attention between the red and remaining patches is highly fragmented, with many small entries in the attention matrix. Even though STA includes vastly more connections to different patches, these only amount to 74.2\% of the total attention, as opposed to 87.3\% for the IA model. Hence, the amount of attention per connection is much smaller for STA, i.e. spatial attention is much more fragmented. This leads to a substantial degradation of accuracy for STA, which as shown on the right of Figure~\ref{fig:portion} is even lower than that of IA. The supplementary presents a more extensive discussion of these issues. 

%and propose a novel {\it task attention block\/} (TAB) to implement {\it cross-task attention\/} (CTA).

\vspace{-8pt}
\paragraph{\bf DNE Model:} To avoid this problem, the proposed DNE model decouples attention into 1) spatial attention, which continues to be performed within each task solely, using~(\ref{eq:sl_transf}) (as in the IA model), and 2) cross-task attention (CTA), which is implemented at the level of the MLP block of (\ref{eq:rl_transf}). As shown at the bottom of figure~\ref{fig:teaser}, the standard MLP mixes the features produced, for each patch, by the different heads of {\it each\/} task expert, processing features of different tasks independently. This is implemented as
\begin{align}
    r_p^i &= s_p^i + \text{FC}(\text{LN}(o_p^i)) \label{eq:MLP1}\\
    o_p^i &= \text{GELU}(\text{FC}(\text{LN}(s_p^i))) \label{eq:MLP2}
\end{align}
where $p\in\{0,1,\dots,P-1\}$ is the patch index, $i\in\{1,2,\dots,t\}$ the task index, FC is a fully connected linear layer and GELU the Gaussian Error Linear Unit, and $o_p^i$ is the intermediate output for $p$-th path of task $i$ features. 

We propose to simply complement this by the yellow CTA connections of  the top model of Figure~\ref{fig:teaser}. This transforms the MLP into a {\it task attention block\/} (TAB). Assume that the expert of task $t$ has $H_t$ spatial attention heads. The input of the TA block is
\begin{equation}
    s_p^t = s_p^{t,1}\oplus s_p^{t,2}\oplus\dots\oplus s_p^{t,H_t}
\end{equation}
where $s_p^{t,i}$ is the tensor output for patch $p$ of spatial attention head $i$ of task expert $t$. DNE does not change the outputs $r^1, \dots, r^{t-1}$ nor the intermediate responses $o^1, \dots, o^{t-1}$ of the TA blocks of the experts previously learned, which are frozen at step $t$. The only operation used to implement (\ref{eq:rlt}) is, for the task $t$ expert, (\ref{eq:MLP1})-(\ref{eq:MLP2}) are replaced by
\begin{align}
    r_p^t &= s_p^t + \text{TA}(\text{LN}(o_p^1, o_p^2, \dots, o_p^t)) \label{eq:TAB1}\\
    o_p^t &= \text{GELU}(\text{TA}(\text{LN}(s_p^1, s_p^2, \dots, s_p^t))), \label{eq:TAB2}
\end{align}
where TA is the {\it task attention\/} operation.

\begin{figure*}[htbp]
    \centering
    \includegraphics[width=0.8\linewidth]{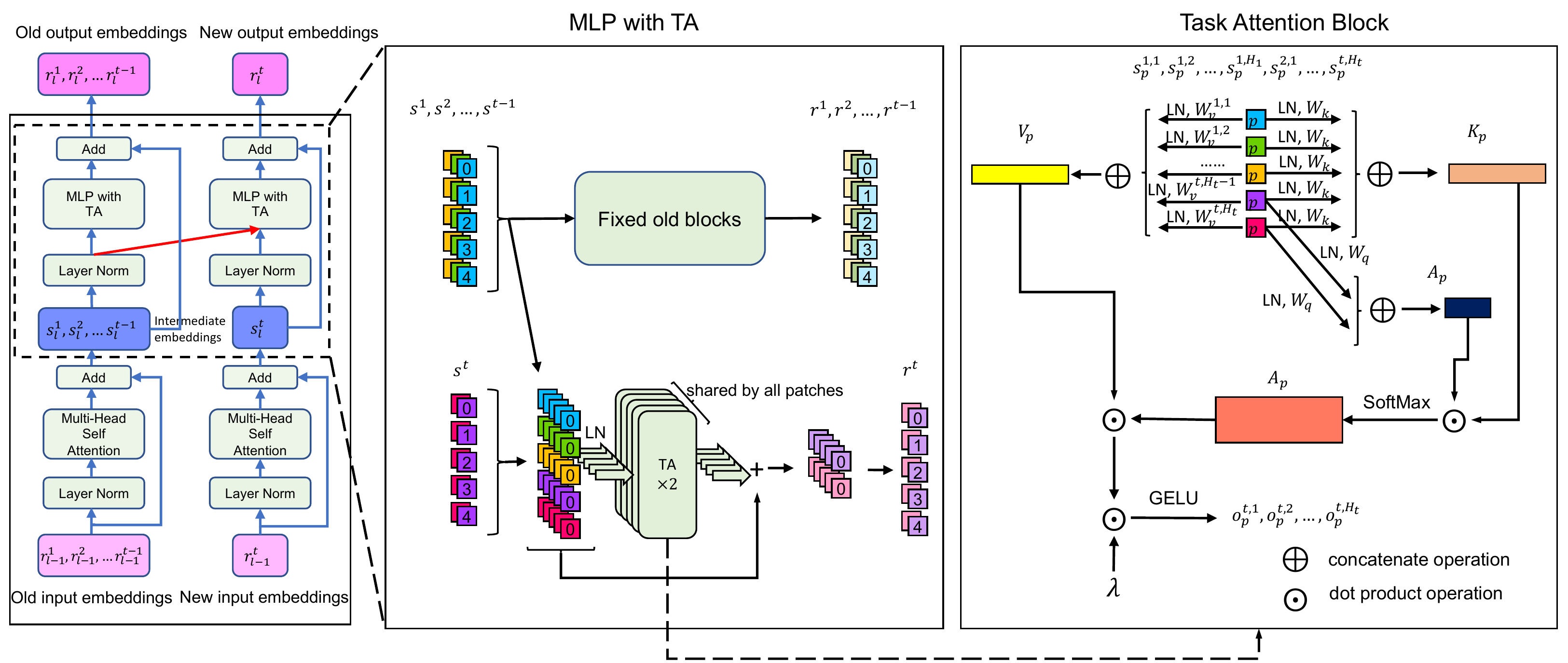}
    \caption{Illustration of the Cross-Task Attention. Left: Overview of the transformer block. Middle: MLP layer with Task Attention Block. Right: Architecture of Task Attention Block. Layer index $l$ is omitted in the middle and right for simplicity. Patch embedding is divided into multiple heads which belong to different tasks. Heads of old tasks (old heads) are handled with fixed old blocks. Correlations between new heads and old heads (cross-task attention) and new heads (self attention) are computed through key matrix $K_p$, query matrix $Q_p$ and their softmaxed dot product, the attention $A_p$. Each input head is projected with an individual value matrix $W_v^{i,j}$ and weighted summed by $A_p$ to generate the output. $\lambda$ is the scaling factor.}
    \label{fig:mlp}
\end{figure*}

Figure~\ref{fig:mlp} illustrates how the proposed TA block fuses features of different heads. 
Consider~(\ref{eq:TAB2}). TA is implemented by an attention block that recombines the spatial attention features $\{s^1_p, \ldots, s_p^{t-1}\}$ generated by previous task experts for patch $p$ with those ($s_p^t$) of the current expert. 
For this, the features $s_p^t$ of task $t$  form a {\it query matrix}
\begin{align}
    &Q_p = [Q_p^{t, 1}, Q_p^{t, 2}, \dots, Q_p^{t, H_t}] \in \mathbb{R}^{D\times H_t}, \mbox{where} \nonumber \\
    &Q_p^{t,i} = W_q\text{LN}(s_p^{t, i}) \in \mathbb{R}^D
    \label{eq:Qp}
\end{align}
is the query vector for the $i$-th head, $p$-th patch in $s^t$, and $W_q \in \mathbb{R}^{D\times D}$ is a learned matrix. 
%and $D$ the depth of each patch tensor. 
A {\it key matrix\/}
\begin{align}
     &K_p = [K_p^{1, 1}, \dots, K_p^{1, H_1}, K_p^{2, 1}, \dots, K_p^{t, H_t}] \in \mathbb{R}^{D\times H} \nonumber \\
     &K_p^{i, j} = W_k\text{LN}(s_p^{i, j}) \in \mathbb{R}^{D}
\end{align}
where $K_p^{i, j}$ is the key vector for the $j$-th head $p$-th patch in $s^i$, $W_k \in \mathbb{R}^{D\times D}$ a learned matrix, and $H=\sum_{k=1}^tH_k$ the total number of heads, is then created from the features produced, for patch $p$, by {\it all\/} task experts. The similarity between query and key features is then captured by the dot-product matrix
$C_p = Q_p^TK_p\in \mathbb{R}^{H_t\times H},$
which is normalized, with a per-row softmax, to obtain the attention weight matrix $A_p$, of rows
\begin{equation}
    A_p^i = \text{SoftMax}(C_p^i/\sqrt{D})\in \mathbb{R}^H
\end{equation}
where $C_p^i$ is the $i$-th row of $C_p$. Row $A_p^i$ contains the set of $H$ weights that determine the relevance of the features generated by each of the spatial attention heads in the model to the features computed by the $i^{th}$ head of task $t$.

The \textit{value matrix} $V_p$ is similar to the \textit{key matrix},
\begin{align}
     &V_p = [V_p^{1, 1}, \dots, V_p^{1, H_1}, V_p^{2, 1}, \dots, V_p^{t, H_t}] \in \mathbb{R}^{D'\times H} \nonumber \\
     &V_p^{i, j} = W_v^{i,j}\text{LN}(s_p^{i, j}) \in \mathbb{R}^{D'}
\end{align}
where $V_p^{i, j}$ is the value vector for the $j$-th head, $p^{th}$ patch in $s^i$, $D'=\gamma D$ the dimension of intermediate heads with an expansion factor $\gamma$ and $W_v^{i,j} \in \mathbb{R}^{D'\times D}$ a learned matrix. Note that, for the \textit{query} and \textit{key}, the matrices $W_q$ and $W_k$ are shared by all heads. However, a separate $W_v^{i,j}$ is used per head for the \textit{value matrix}. This is because the value vectors are the components of the TA block output. Since different heads encode knowledge of different tasks, the matrix $W_v$ can be seen as translating knowledge across task domains. This benefits from the added flexibility of a matrix per task. 

Finally, the $i$-th head of output $o_p^t$ is
\begin{equation}
    o_p^{t, i} = \text{GELU}(\lambda_i\sum_{j=1}^H A_p^{i, j}V_p^j)\in \mathbb{R}^{D'}
    \label{eq:DNEout}
\end{equation}
 where $A_p^{i,j}$ is the $j$-th element of $A_p^i$, $V_p^j$ the $j$-th column of $V_p$, and $\lambda_i$ a learned scalar. 
 The procedure used to implement (\ref{eq:TAB1}) is identical to (\ref{eq:Qp})-(\ref{eq:DNEout}), but the dimension of each head changes from $D'$ to $D$. 
 Note that the entries $A_p^{i,j}$ measure the similarity between the query $Q_p^{t,i}$ generated by the $i^{th}$ head of task expert $t$ for patch $p$ and the keys $K_p^j$ generated for the patch by each of the $H = \sum_{k=1}^t H_k$ heads of all task experts, as illustrated by the yellow connections of Figure~\ref{fig:teaser}. If $A_p^{ij} = 1, \forall i,j$, the TA block simplifies to a generalization of the MLP of~(\ref{eq:rl_transf}), whose linear layers implement projections from all tasks to the current one. 

\vspace{-10pt}
\paragraph{\bf Training Objective:}
The training objective of DNE is the weighted average of three losses. Task $t$ is trained with 1) a cross entropy classification loss $\mathcal{L}_{ce}$ defined over the class set of all tasks, applied at the output of the classifier $g$ of (\ref{eq:g}), 2) a task expertise loss $\mathcal{L}_{te}$ that encourages task expert $t$ to learn features that are more informative of the new task,  and 3) a distillation loss $\mathcal{L}_{dis}$ that assures that the expert has good performance on the classes of the previous tasks. The task expertise loss was first introduced by DER\cite{der}. It treats all previous classes  as one ($\mathcal{Y}' = \bigcup_{i=1}^{t-1} \mathcal{Y}_i$) and conducts a $|\mathcal{Y}_t|+1$-way classification. The distillation loss is a KL divergence between the common outputs of $g$ after tasks $t-1$ and $t$. This is implemented by applying a softmax to the logits of classes in $\mathcal{Y}' = \bigcup_{i=1}^{t-1} \mathcal{Y}_i$ of $g^{t-1}$ and $g^t$, and computing the KL divergence between the two distributions. 
%after a SoftMax layer.{\color{red} WHAT IS THIS? A KL divergence is between probability distribitions, not logits, and $g$ has different numbers of classes at $t-1$ and $t$, take the output logits of the old model and the new model as input} {\color{blue} Sorry, I should make it more clear, the KL divergence is used over the softmaxed logits and is between the old classes, i.e. $g^{t-1}$ and $g^t[:\sum_{i=1}^{t-1} |\mathcal{Y}_i|]$}. 
Minimizing this KL divergence guarantees that task expert $t$ performs similarly to task expert $t-1$ on the classes of all previous tasks. 
Standard procedures~\cite{decouple} are used to account for the data imbalance between the current task and memory buffer data.
%{\color{red} DON'T YOU NEED TO TALK ABOUT THE BUFFER EXAMPLES AND CLASS-BALANCED SAMPLING, ETC.? OR MAYBE SAY THAT THIS IS DONE AS IN SOME OTHER METHOD AND THEN GIVE THE DETAILS IN APPENDIX?}

\paragraph{Discussion:} DNE leverages the fact that the MLP plays a critical role in the creation of new features by the transformer. Like any $1 \times 1$ convolution operator, its main function is to perform the feature mixing that transforms features of lower semantic abstraction at the bottom of the network into the features of high-level semantics at the top. Since DNE aims to {\it reuse\/} existing features and {\it combine\/} them with the information of the new task to {\it create\/} new features that account for the information not already captured by the CIL model, the MLP block is naturally suited to implement CTA. The proposed TAB simply extends feature mixing across tasks. Hence, the DNE model can reuse ``old" knowledge to solve new tasks. Rather than having to relearn all features by itself, task $t$ expert inherits the features already computed by the experts of the previous tasks. 

This is similar to distillation approaches to CIL, which use the features produced by the previous models to regularize the features of task $t$. However, in distillation, the model is replicated and produces a new feature vector that is constrained to be similar to that of the existing model. In DNE, because the new task expert {\it reuses\/} the features computed by the previous experts {\it throughout the network\/}, it can be a small network. In result, the model is mostly frozen and only a small branch added per task, enabling a better trade-off between size and accuracy. This also provides DNE with a better trade-off between accuracy and complexity than the IA model of NE, which repeats the full network per task.

On the other hand, by implementing spatial attention with~(\ref{eq:sl_transf}) and CTA with (\ref{eq:TAB1})-(\ref{eq:TAB2}), DNE is immune to the fragmentation of attention of the STA model. Note that, with respect to Figure~\ref{fig:fourtype}, spatial attention is implemented exactly as in the IA model: it relies uniquely on the dot-products of red and green patches. CTA is then performed, using (\ref{eq:TAB1})-(\ref{eq:TAB2}), once the information  fusion of spatial attention has been accomplished. At this point, the feature projections of the yellow patches already account for all information in the blue patches and attention only relies on dot-products between red and yellow patches.  This prevents the fragmentation of attention that plagues STA.

%We propose to do this using attention, as illustrated in Figure~\ref{fig:teaser}. As illustrated at the top of Figure~\ref{fig:teaser}, DNE instead implements a {\it task attention\/} (TA) block that replaces each of the FC layers by an attention module that mixes features produced, per patch, by the spatial attention heads of {\it all} task experts. As usual for attention, the features are combined according to weights derived from a query tensor, which is now the patch tensor for the current task. The features of this tensor are mixed with those of the patch tensors of the remaining attention heads, using weights derived by feature similarity, as discussed in detail below. 

\noindent{\bf Computation:}
Under the IA model of NE, the spatial attention computation of (\ref{eq:sl_transf}) has complexity $O(P^2D^2)$ per attention head, and the MLP of~(\ref{eq:rl_transf}) computation $O(H^2D^2)$ per patch, where $D$ is the patch token dimension and $H$ the number of heads.  A CIL model of $T$ task experts and $H$ attention heads per expert has computation $O(THP^2D^2 + TPH^2D^2) = O(THPD^2(P+H))$. 
 Under DNE, the complexity of spatial attention is the same, but the current task expert queries the previous task experts for the features generated per patch, using feature similarity to determine the relevance of the old knowledge to the new task. This operation has complexity $O(T^2H^2PD^2)$, for a total complexity of $O(THP^2D^2 + T^2H^2PD^2) = O(THPD^2(P+TH))$. Hence, in principle, DNE has more computation. However, the reuse of features allows a small number of heads $H$ per task. In our implementation, we use $H=1$, which is shown to suffice for good accuracy in the next section. This makes the effective computation of DNE $O(TPD^2(P+T))$. Hence the compute ratio between DNE and DE is $(P+T)/H(P+H)$, and DNE is more efficient if $T < H^2 + (H-1) P$. For the standard configuration of the ViT transformer ($H=12, P=196$), this bound is $T < 2,300$.

\vspace{-10pt}
\section{Experiments}
\label{sec:exp}

In this section, we discuss various experiments conducted to evaluate the CIL performance of DNE.
\vspace{-10pt}
\paragraph{Experimental Setup:} Various experiments were performed with the following set-up.

\textbf{Benchmarks.} All experiments used CIFAR100\cite{cifar} or ImageNet100\cite{imagenet}\footnote{All datasets used in the paper were solely downloaded by the university.}, a first task with $50$ classes and  $N_s$ classes per subsequent task. We denote $N_s$ as the step size.

\textbf{Methods.} \textbf{DNE} was compared to multiple baselines. \textbf{Joint} trains all classes simultaneously. It is not a CIL method but an upper bound. \textbf{Dytox}\cite{dytox} is a transformer based method, using a single network with trained task tokens to generate different feature vectors per task. \textbf{DER}\cite{der} is a NE method, learning a sub-network per task and concatenating the features of all sub-networks for classification. \textbf{FOSTER}\cite{foster} is similar to DER, adding a network per task. However, the new and old networks are distilled into one compressed network to keep the total model size fixed. \textbf{iCaRL}\cite{icarl} and \textbf{PODNet}\cite{podnet} are distillation based methods. They introduce constraints on the logits of pooled intermediate features of the old and new networks.

\textbf{Evaluation metrics.} Let $M$ be the number of tasks and $A_i$ the average accuracy of all known classes after learning task $i$. Performance is measured by \textit{Last Accuracy} $LA=A_M$ and \textit{Average Incremental Accuracy} $AA=\frac{1}{M}\sum_{i=1}^MA_i$. To eliminate the effects of backbones (e.g. DER uses ResNet18\cite{resnet} and Dytox a transformer), we also consider the \textit{Difference} of LA between joint and CIL model, $D = A_{M, joint} - A_{M, model}$.  Finally, methods are compared by \textit{floating point operations per second (FLOPs)} $F$.

\textbf{Implementation details.} DNE and Dytox use a 6-layer Vision Transformer\cite{vit} backbone. Patch size is 4 (16) on CIFAR100 (ImageNet100). DNE learns a 12-head transformer in the first task and adds $k\in\{1,2,4\}$ heads per subsequent task. DER, FOSTER, iCaRL and PODNet use a modified ResNet18\cite{der} (standard ResNet18\cite{resnet}) backbone on CIFAR100 (ImageNet100). A buffer memory $\mathcal{M}$ of $2,000$ examples is used on both datasets. We first train the backbone and classifier with all the data from the current dataset and memory. Similar to \cite{der}, we then build a class-balanced dataset by subsampling the current dataset to tune the classifier. More details are given in the supplementary.

%\begin{table}[t]
%\setlength{\tabcolsep}{2pt}    
%    \centering
%       \scriptsize
%    \begin{tabular}{c|cccc|cccc}
%        \toprule
%        \multirow{2}{*}{Method} & \multicolumn{4}{c|}{CIFAR100, $N_s$=10} & \multicolumn{4}{c}{ImageNet100, $N_s$=10}\\
%        \cline{2-9}
%        & $LA\uparrow$ & $AA\uparrow$ & $D\downarrow$ & $S\downarrow$ & $LA\uparrow$ & $AA\uparrow$ & $D\downarrow$ & %$S\downarrow$ \\
%        \midrule
%        Joint (Transf.)  & 76.12 & - & 0 & 41M & 79.12 & - & 0 & 41M\\
%        Joint (ResNet18)     & 80.41 & - & 0 & 43M & 81.20 & - & 0 & 43M\\
%        \midrule
%        iCaRL               & 44.72 & 59.32 & 31.40 & 43M & 42.84 & 55.65 & 38.36 & 43M\\
%        PODNet              & 52.46 & 66.41 & 27.95 & 45M & 63.46 & 73.57 & 17.74 & 45M\\
%        DER                 & 63.78 & 71.69 & 16.63 & 258M & 70.40 & 76.90 & 10.80 & 258M\\
%        FOSTER              & 63.31 & 72.20 & 17.10 & 44M & 67.68 & 75.85 & 13.52 & 44M\\
%        Dytox               & 64.06 & 71.55 & 12.06 & 41M& 68.84 & 75.54 & 10.28 & 46M\\
%        \midrule
%        DNE-1head     & 68.04 & 73.68 &  8.08 & 62M & 72.30 & 78.09 &  6.82 & 64M\\
%        DNE-2heads    & 69.73 & 74.61 &  6.39 & 70M & \textbf{73.64} & \textbf{78.88} &   \textbf{5.48} & 72M\\
%        DNE-4heads    & \textbf{70.04} & \textbf{74.86} &  \textbf{6.08} & 93M & 73.58 & 78.56 &   5.54 & 96M\\
%        \bottomrule
%    \end{tabular}
%    \caption{Comparison of CIL approaches on CIFAR100 and ImageNet100, for $N_s = 10$.}
%    \label{tab:comp1}
%\end{table}

\begin{table}[t]
\setlength{\tabcolsep}{2pt}    
    \centering
       \scriptsize
    \begin{tabular}{c|cccc|cccc}
        \toprule
        \multirow{2}{*}{Method} & \multicolumn{4}{c|}{CIFAR100, $N_s$=10} & \multicolumn{4}{c}{ImageNet100, $N_s$=10}\\
        \cline{2-9}
        & $LA\uparrow$ & $AA\uparrow$ & $D\downarrow$ & $F\downarrow$ & $LA\uparrow$ & $AA\uparrow$ & $D\downarrow$ & $F\downarrow$ \\
        \midrule
        Joint (Transf.)  & 76.12 & - & 0 & 1.38G & 79.12 & - & 0 & 1.38G\\
        Joint (ResNet18)     & 80.41 & - & 0 & 1.12G & 81.20 & - & 0 & 1.12G\\
        \midrule
        iCaRL               & 44.72 & 59.32 & 31.40 & 1.12G & 42.84 & 55.65 & 38.36 & 1.12G\\
        PODNet              & 52.46 & 66.41 & 27.95 & 1.12G & 63.46 & 73.57 & 17.74 & 1.12G\\
        DER                 & 63.78 & 71.69 & 16.63 & 6.68G & 70.40 & 76.90 & 10.80 & 6.68G\\
        FOSTER              & 63.31 & 72.20 & 17.10 & 1.12G & 67.68 & 75.85 & 13.52 & 1.12G\\
        Dytox               & 64.06 & 71.55 & 12.06 & 1.38G& 68.84 & 75.54 & 10.28 & 1.38G\\
        \midrule
        DNE-1head     & 68.04 & 73.68 &  8.08 & 2.68G & 72.30 & 78.09 &  6.82 & 2.68G\\
        DNE-2heads    & 69.73 & 74.61 &  6.39 & 3.10G & \textbf{73.64} & \textbf{78.88} &   \textbf{5.48} & 3.10G\\
        DNE-4heads    & \textbf{70.04} & \textbf{74.86} &  \textbf{6.08} & 4.02G & 73.58 & 78.56 &   5.54 & 4.02G\\
        \bottomrule
    \end{tabular}
    \caption{Comparison of CIL approaches on CIFAR100 and ImageNet100, for $N_s = 10$.}
    \label{tab:comp1_flop}
\end{table}
\begin{table}[t]
\setlength{\tabcolsep}{2pt}    
    \centering
       \scriptsize
    \begin{tabular}{c|cccc|cccc}
        \toprule
        \multirow{2}{*}{Method} & \multicolumn{4}{c|}{CIFAR100, $N_s$=5} & \multicolumn{4}{c}{CIFAR100, $N_s$=25}\\
        \cline{2-9}
        & $LA\uparrow$ & $AA\uparrow$ & $D\downarrow$ & $F\downarrow$ & $LA\uparrow$ & $AA\uparrow$ & $D\downarrow$ & $F\downarrow$ \\
        \midrule
        Joint(Transf.)  & 76.12 & - & 0 & 1.38G & 76.12 & - & 0 & 1.38G\\
        Joint(ResNet18)     & 80.41 & - & 0 & 1.12G & 80.41 & - & 0 & 1.12G\\
        \midrule
        iCaRL               & 42.89 & 55.23 & 33.23 & 1.12G & 53.79 & 66.80 & 26.62 & 1.12G\\
        PODNet              & 48.18 & 60.69 & 27.94 & 1.12G & 61.54 & 71.45 & 18.87 & 1.12G\\
        DER                 & 60.73 & 70.42 & 15.39 & 12.19G & 69.08 & 74.82 & 11.33 & 3.32G\\
        FOSTER              & 48.18 & 60.69 & 27.94 & 1.12G & 70.14 & \textbf{76.33} & 10.27 & 1.12G\\
        Dytox               & 58.59 & 68.31 & 17.53 & 1.38G & 69.29 & 74.10 & 6.83 & 1.38G\\
        \midrule
        DNE-1head    & 69.10 & 74.03 &  7.02 & 5.71G & 67.61 & 73.27 & 8.51 & 1.19G\\
        DNE-2heads    & \textbf{69.72} & \textbf{74.27} &  \textbf{6.40} & 7.39G & 68.99 & 73.91 &  7.13 & 1.27G\\
        DNE-4heads    & 69.43 & 74.20 &  6.69 & 10.75G & \textbf{71.47} & 75.70 & \textbf{4.65} & 1.45G\\
        \bottomrule
    \end{tabular}
    \caption{Comparison of CIL methods on CIFAR100 for different step sizes.}
    \label{tab:comp2_flop}
\end{table}

\begin{figure*}\RawFloats
\begin{minipage}{\linewidth}
\centering
\footnotesize
\setlength{\tabcolsep}{2pt}
\begin{tabular}{cccccc}
    \includegraphics[width=0.16\linewidth]{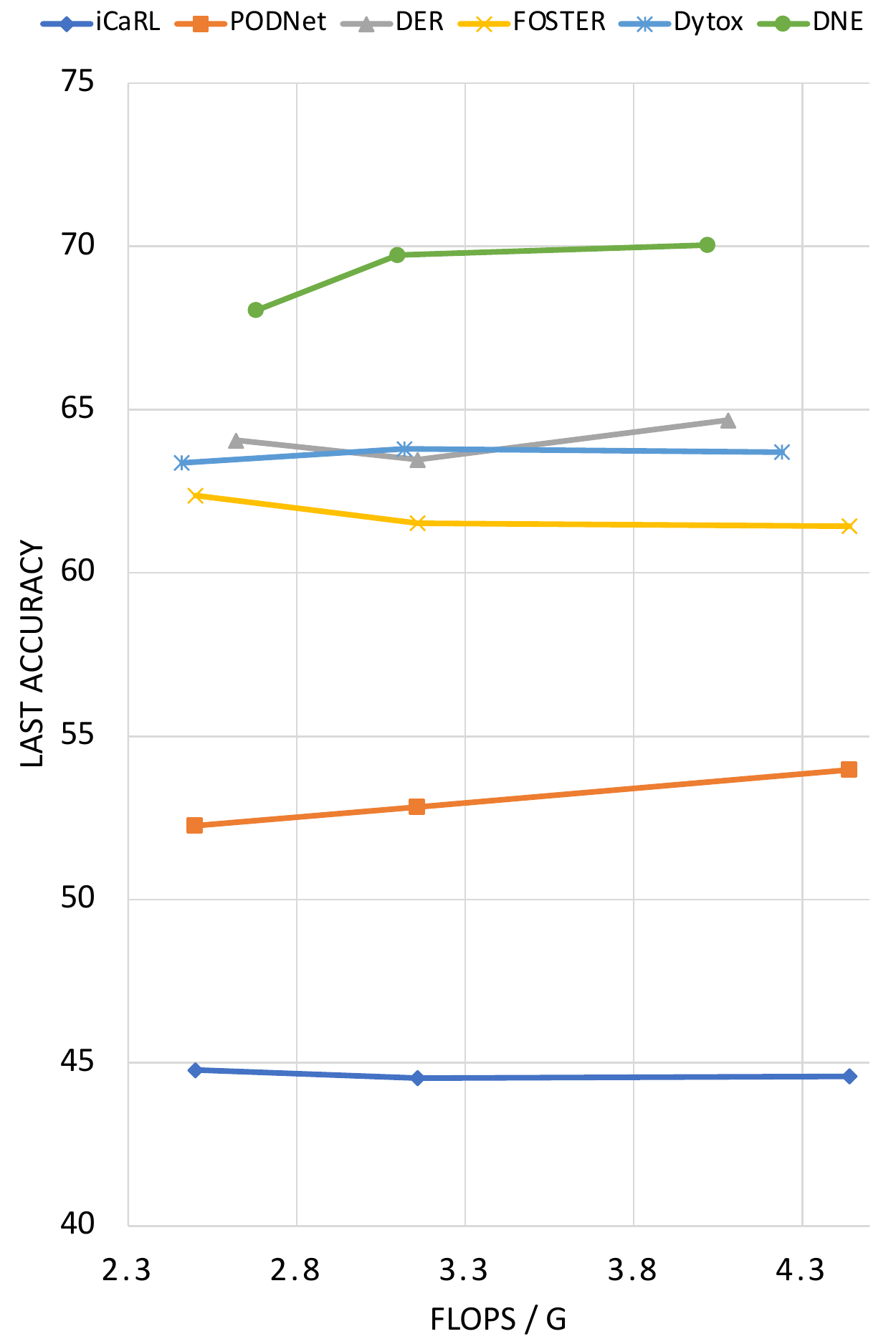} &
    \includegraphics[width=0.16\linewidth]{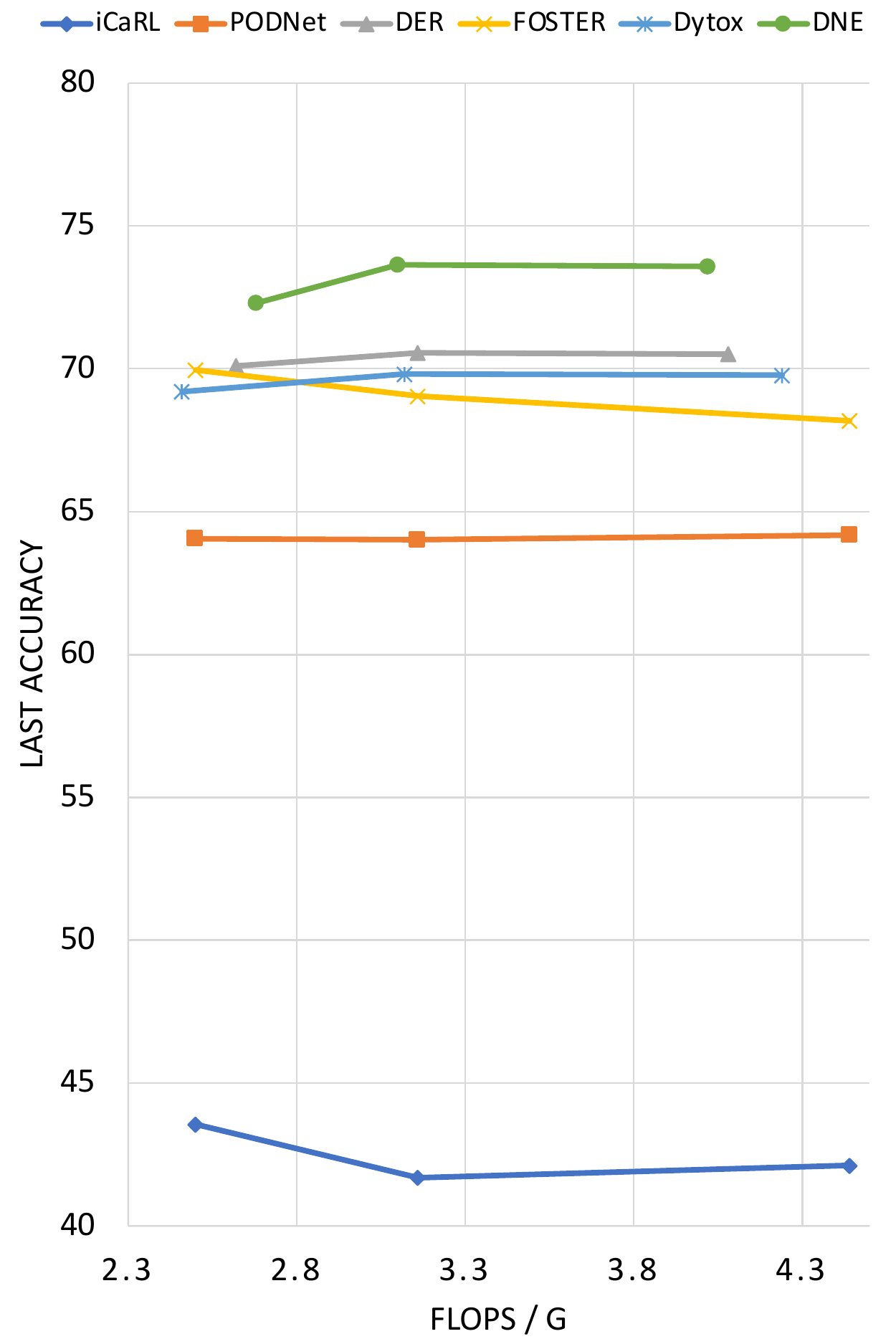} &
    \includegraphics[width=0.16\linewidth]{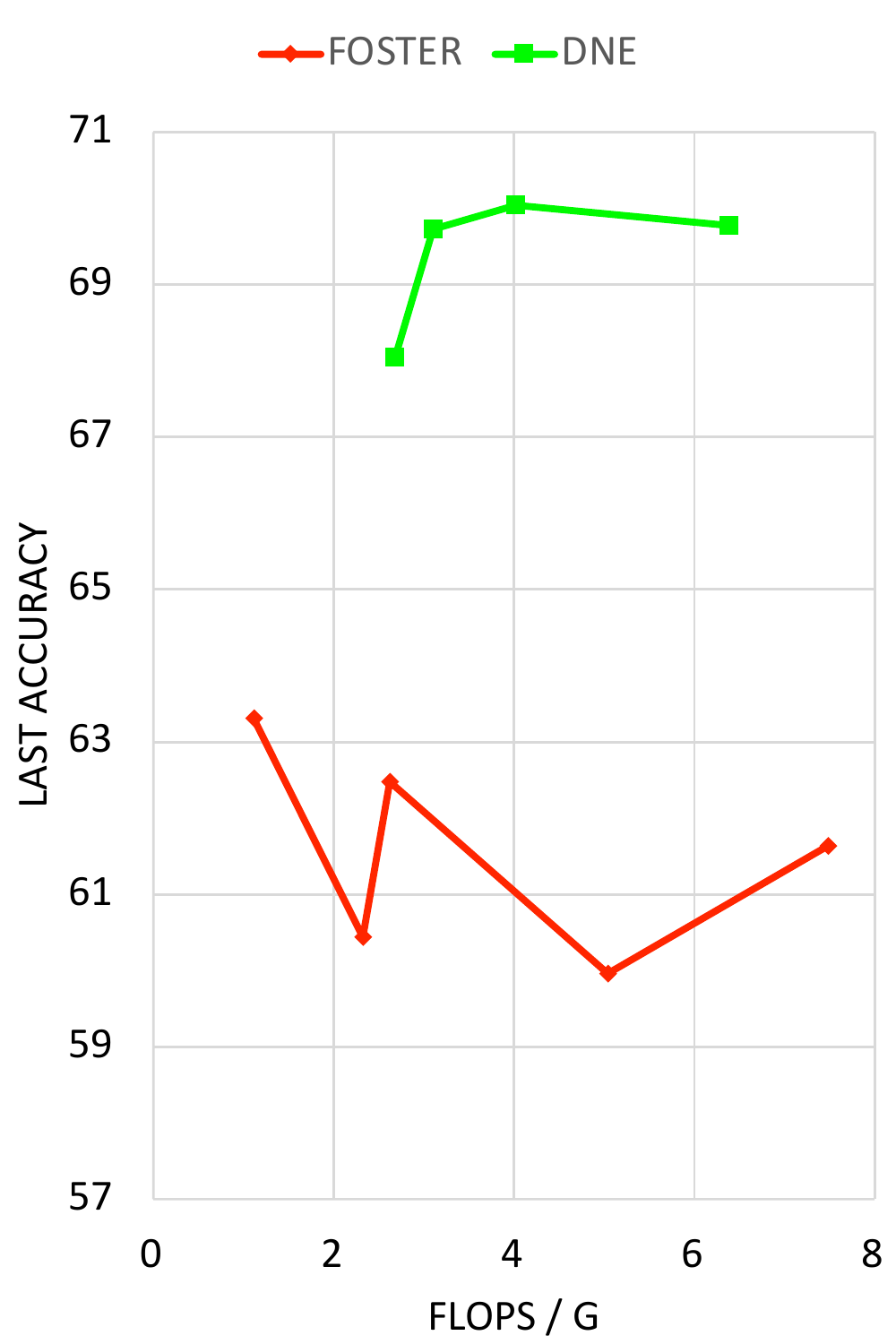} &
    \includegraphics[width=0.16\linewidth]{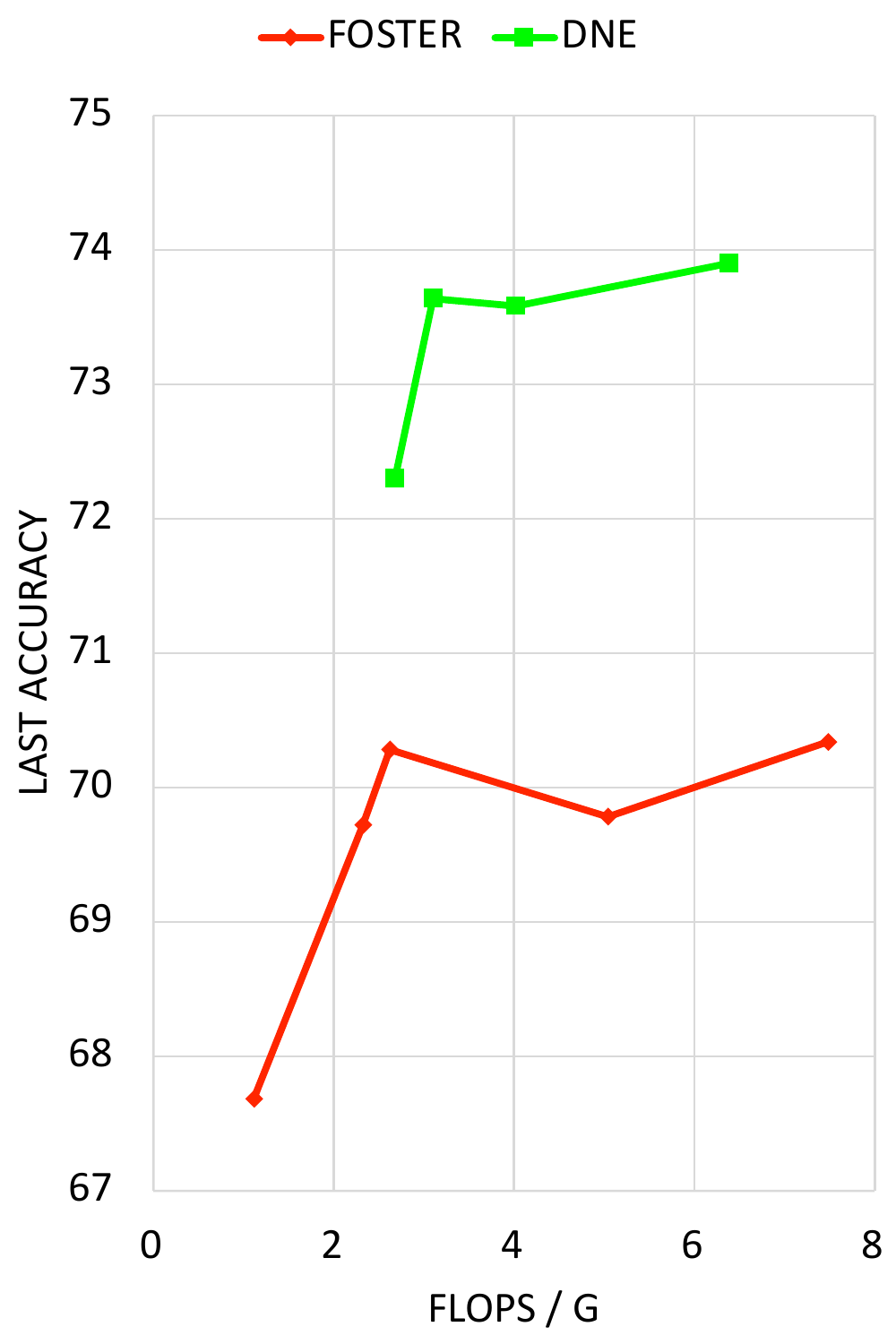} &
    \includegraphics[width=0.16\linewidth]{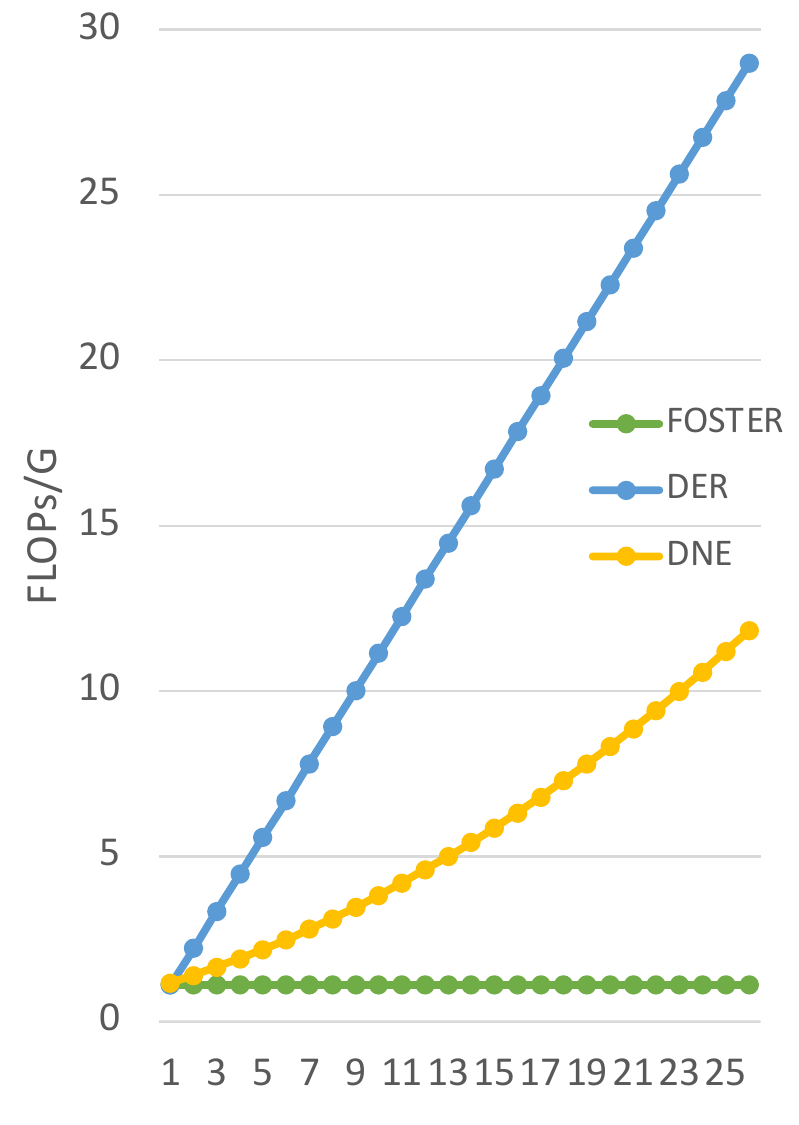} &
    \includegraphics[width=0.16\linewidth]{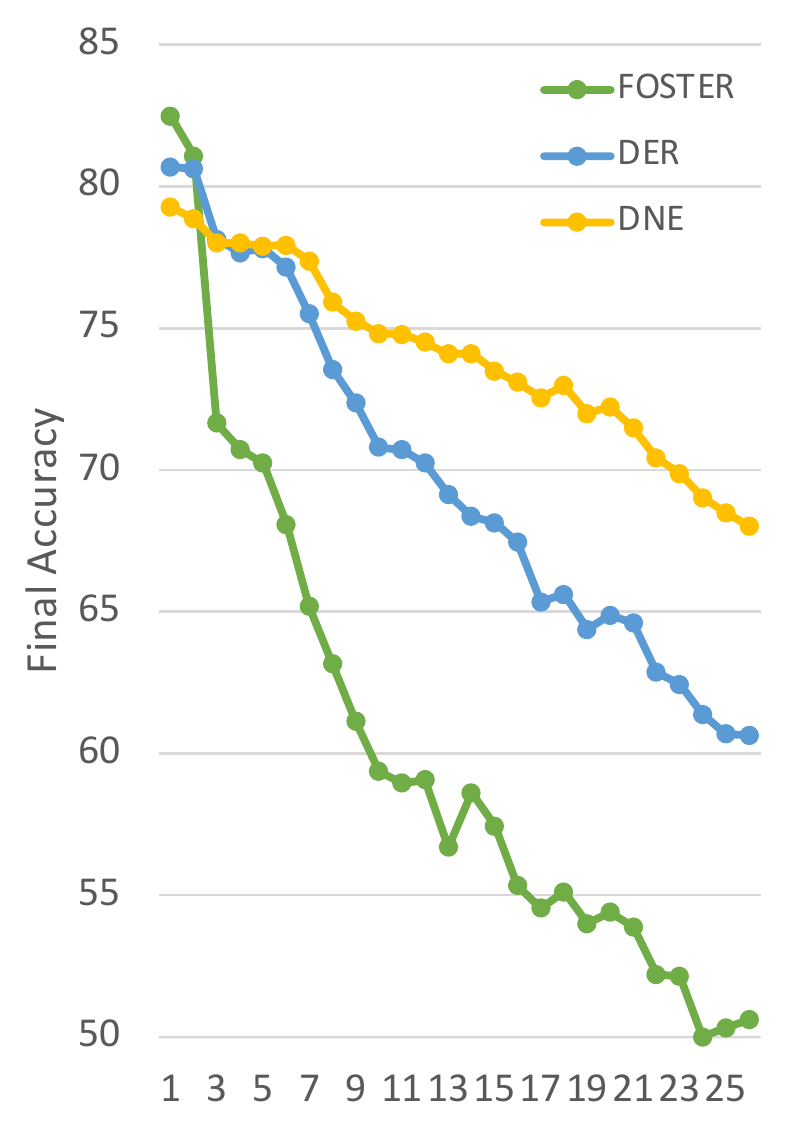} \\
    \scriptsize{(a)} & \scriptsize{(b)} & \scriptsize{(c)} & \scriptsize{(d)} & \scriptsize{(e)} & \scriptsize{(f)}
\end{tabular}
\caption{Accuarcy-Scale Trade-off: (a)(b)FLOPs-Accuracy Trade-off, on CIFAR100/ImageNet100 $N_s=10$, (c)(d) Comparison to FOSTER with different backbones, on CIFAR100/ImageNet100 $N_s=10$, (e)(f) FLOPs/Acc vs task step, on CIFAR100 $N_s=2$}
\label{fig:tradeoff}
\end{minipage}

%\begin{minipage}{.35\linewidth}
%\centering
%\footnotesize
%\setlength{\tabcolsep}{2pt}
%\begin{tabular}{ccc|c}
%    \includegraphics[width=0.49\linewidth]{imgs/foster_a.pdf} &
%    \includegraphics[width=0.49\linewidth]{imgs/foster_b.pdf} \\
%     CIFAR100 & ImageNet100
%\end{tabular}
%\caption{Comparison to FOSTER with different backbones, $N_s=10$.}
%\label{fig:foster}
%\end{minipage}
\end{figure*}

\vspace{-15pt}
\paragraph{Accuracy Comparison to SOTA:}
Table~\ref{tab:comp1_flop} compares DNE to SOTA methods, for $N_s = 10$.  On both datasets, DNE outperforms all methods by a clear margin. On CIFAR100, even with only 1 head per task expert, DNE reaches an LA of 68.04\%, which is 3.98\% higher than Dytox, the current SOTA. Regarding AA, DNE-1head is 1.48\% higher than the SOTA, FOSTER.  On ImageNet100, DNE-1head outperforms the SOTA, DER, by 1.9\% in LA and 1.19\% in AA.   With more heads, DNE performs even better. Between 1 and 4 heads, LA increases by 2\% (1.28\%) on CIFAR100 (ImageNet100). On ImageNet100, performance saturates with 2 heads per task, due to the increased image resolution and number of patches, which allow the learning of more discriminant features and more efficient knowledge sharing between tasks.

DNE gains are even more significant for the difference $D$ to joint model. While the transformer of DNE has 2-4\% weaker joint accuracy than the ResNet18 of DER or FOSTER, DNE significantly outperforms the latter, because its difference $D$ is $2$ to $3$ times smaller. When compared to the SOTA Dytox model, DNE reduces $D$ by as much as 50\%. These results illustrate the effectiveness of the TA module of Figure~\ref{fig:mlp} for sharing information across tasks. 

These conclusions hold qualitatively across step sizes $N_s$. Table~\ref{tab:comp2_flop} shows results for $N_s$ as different as $5$ and $25$. The advantages of DNE are larger for smaller step sizes. When $N_s=5$, DNE-2heads outperforms the SOTA, DER, by 8.99\% in LA. For $N_s=25$, the LA of DNE-4heads is 1.33\% higher than that of the SOTA, FOSTER. This is most likely because DNE freezes the old network - only the added task experts can learn about new tasks. Hence, the number of extra heads determines the model capacity for learning. When $N_s$ is small, 1head is enough to handle the new task classes. However, as the number of classes per task ($N_s$) grows, larger capacity is eventually needed. This explains the increase in accuracy from the 1head to the 4head model when $N_s = 25$.
However, as the number of heads increases, DNE becomes more similar to DER or FOSTER, which uses a full network per task. Hence, the benefits of its feature sharing across tasks are smaller.

\vspace{-15pt}
\paragraph{Accuracy-Scale Trade-off:}
A criticism of NE is the network growth with the number of tasks. However, growth is inevitable in CIL, since the ever-growing task sequence will eventually exceed the capacity of any fixed model. The real question is not \textit{whether the network needs to grow} but \textit{how fast it needs to grow}. For insight on this, we compared the accuracy-scale trade-off of the different methods. 

We started by changing the dimension of the feature vector of all methods to match their model sizes with those of DNE for $k\in\{1,2,4\}$ extra heads. Experiments were conducted on CIFAR100 and ImageNet100 with $N_s=10$. Detailed configurations are listed in the supplementary materials. Figure~\ref{fig:tradeoff}(a)(b) compares the accuracy-scale trade-off of the different methods. It is clear that the accuracy of single model methods (iCaRL, PODNet, FOSTER, Dytox) fails to improve with model size. This is unsurprising, since these methods use the same network to solve all tasks. Since the network must have enough capacity to solve all tasks, there is little benefit to a wider feature vector or a deeper architecture. This is unlike DNE, whose performance can increase with model size and FLOPs, by leveraging larger task experts as $N_s$ increases. Overall, DNE has the best trade-off on both datasets. 
%In most cases, {\it its LA performance with the smallest model size is SOTA even when compared to the models of the largest size of previous methods.} The only exception is CIFAR100 with $N_s=25$, where small models ($<$55 MB) lack the per-expert capacity to solve each task. It is only in this case that larger task experts are needed to beat the SOTA. The gains of DNE over DER are particularly informative of the benefits of feature sharing and cross-task attention. 
 
% However, in these experiments. accuracy does not increase significantly with model size. This suggests that the datasets do not challenge the capacity of even the smaller models. 
 %The main exceptions are CIFAR100 with $N_s = 25$ and ImageNet 100 with $N_s=10$. {\color{red} WHAT SETTING COULD WE HAVE THAT CHANGED THIS? LARGER FIRST TASK? SMALLER? SOMETHING ELSE?}
%To eliminate this problem, we performed 
A second comparison was performed by changing the network backbone.
For simplicity,  we limited this comparison to FOSTER. Figure~\ref{fig:tradeoff}(c)(d) compares results of  FOSTER with different backbones, from ResNet18 to ResNet152, and DNE with numbers of heads $k\in\{1,2,4,7\}$. Clearly, FOSTER never approaches the trade-off of DNE.

Finally, we consider the growth rate of DNE and other methods. We set $N_s=2$ on CIFAR100, which leads to an experiment with 26 tasks, the maximum task number we can reach. The FLOPs of DNE, DER (NE based method) and FOSTER (Distillation based method), as well as their final accuracy, for different task steps are illustrated in Figure~\ref{fig:tradeoff}(e)(f). FOSTER, which maintains a roughly constant FLOPs, suffers form severe catastrophic forgetting as more tasks are added. DER performs much better than FOSTER, but consumes 26 times the number of FLOPs. As discussed above, the FLOPs of DNE are quadratic on the number of tasks. However, as the dense connections allow feature reuse and a single head per task expert, the actual FLOP growth is relatively slow. Hence, DNE has much lower FLOPs than DER (which has linear FLOPs rate with number of tasks) over a large range of task setps. On the other hand, the performance of DNE is also much better than those of DER and FOSTER. In summary, DNE is both simpler and better than the previous approaches.

\vspace{-10pt}
\section{Conclusion}
\label{sec:end}
In this work, we have pointed out that NE-based CIL methods have unsustainable model growth for most practical applications and reformulated the NE problem, to consider the trade-off between accuracy and model size. We then proposed the DNE approach to address this trade-off, and introduced an implementation of DNE that relies on a new TAB to perform cross-task attention. Experiments have shown that DNE outperforms all previous CIL methods both in terms of accuracy and of the trade-off between accuracy and model size.  

%ystematically analyze the network expansion methods in class incremental learning, based on the observation of previous methods' limitation, we proposed the Dense Network Expansion (DNE) method, which can better transfer knowledge from old classes to new classes and reaches a better accuracy-size trade-off. Extensive experiments show that DNE is effective and outperforms existing SOTA methods by a clear margin.

%\noindent{\bf Limitations:} DNE has a few limitations. First, it only addresses classification. Future extensions should  consider object detection, segmentation, unsupervised learning, etc. Second,  more realistic and larger datasets should be considered to validate DNE. Third, current CIL settings assume a relatively small number of tasks. Like all methods in the literature, it is unclear how DNE can handle problems with hundreds of tasks. Fourth, DNE uses a memory buffer for data of previous tasks. Ideally, this should not be necessary.

\noindent{\bf Acknowledgment:} This work was partially funded by NSF award IIS-2041009, and a gift from Qualcomm. We also acknowledge the Nautilus platform, used for the experiments.

%%%%%%%%% REFERENCES
{\small
\bibliographystyle{ieee_fullname}
\bibliography{egbib}
}

\newpage

\noindent\textbf{\Large Appendix}
\appendix
\renewcommand{\thefigure}{A}
In this supplementary material, we discuss more details about the architecture of the proposed Dense Network Expansion (DNE) method, the experiment setup and add several experiments to further validate the effectiveness of DNE.

%%%%%%%%% BODY TEXT
\section{Model Design}
\subsection{MHSA block}
\begin{figure}[htbp]
    \centering
    \includegraphics[width=\linewidth]{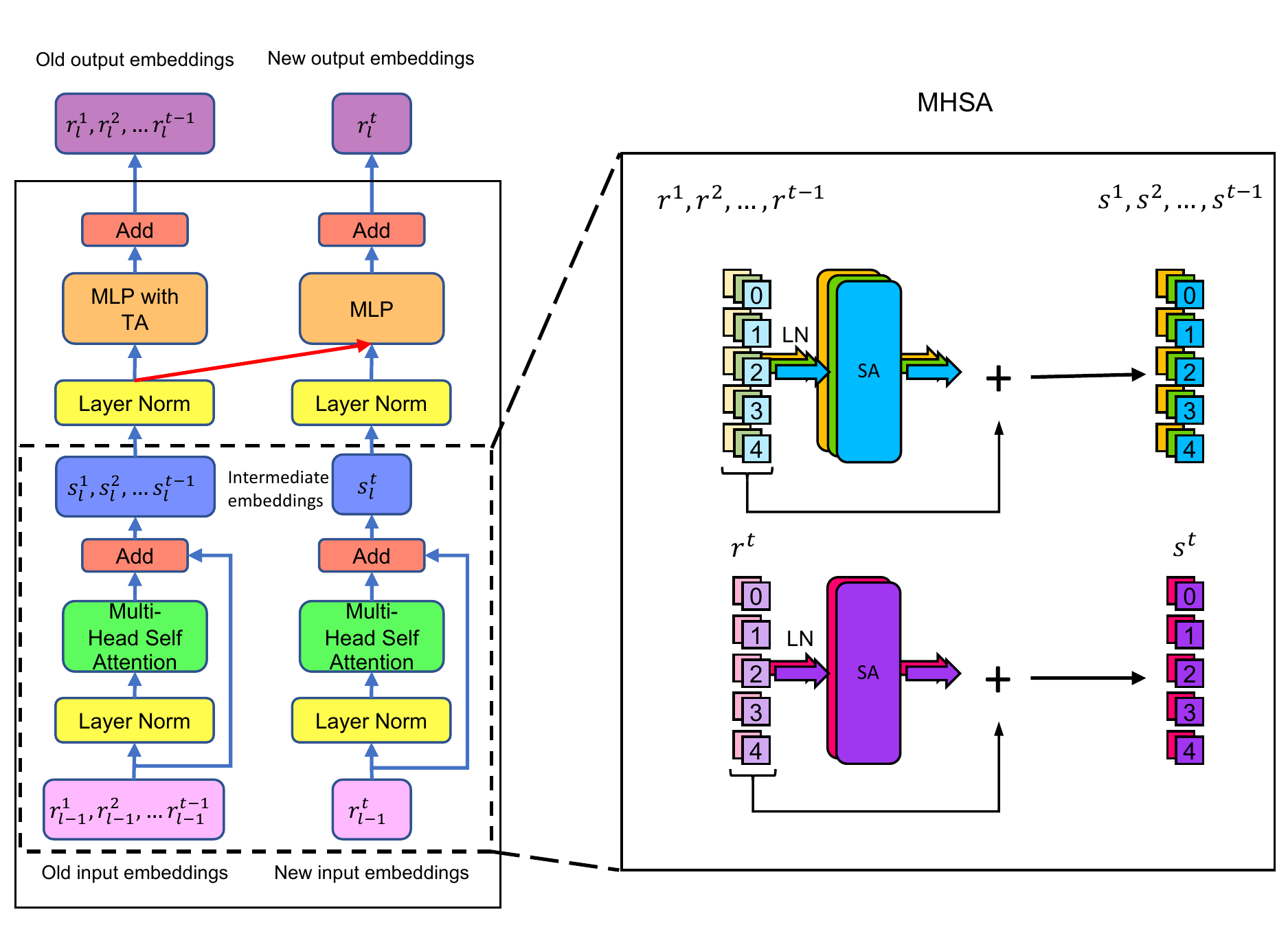}
    \caption{In MHSA, heads belong to old tasks are processed with fixed old self-attention blocks. Only the newly added self-attention blocks are trained.}
    \label{fig:mhsa}
\end{figure}

In DNE, cross-task attention is only used in MLP layers of a transformer block. In MHSA a ``sparse'' connection is used, as is illustrated in Figure~\ref{fig:mhsa}.  Specifically, we split the input $r$ into different heads based on Equation~7 and similar to Equation~15 of main paper, which are then processed as:
\begin{align}
    &u = [u^{1,1}, u^{1,2}, \dots, u^{1, H_1}, u^{2,1}, \dots, u^{t, H_t}]\in\mathbb{R}^{P\times D\times H}\\
    &u^{i,j} = \text{SA}^{i,j}(\text{LN}(r^{i,j}))\in\mathbb{R}^{P\times D}
\end{align}
where SA is the self-attention block as in \cite{vit}, $u^{i,j}$ is the intermediate output for head $j$ in task $i$. Note that, only the self-attention blocks of the current task (i.e. $\text{SA}^{i,j}$ with $i=t$) will be trained. After this, all the heads from the same task are fused for the output $s$:
\begin{align}
    s &= s^1\oplus s^2\oplus\dots\oplus s^t\in\mathbb{R}^{P\times (D\times H)}\\
    s^i &= r^i + \text{FC}^i(u^{i,1}\oplus u^{i,2}\oplus\dots\oplus u^{i,H_i})\in\mathbb{R}^{P\times (D\times H_i)}
\end{align}
where FC is a linear layer and $s^i$ the output embedding for $i$-th task.

\subsection{Task tokens}
Transformer-based CIL methods\cite{dytox, lvt, l2p, dual}, typically include an extra block $f_{L+1}$, where global learned task tokens $e = [e^1, e^2, \dots, e^t]$ are used to produce task-specified features and are then taken as the inputs of the classification layer, according to
\begin{eqnarray}
    e' &= & f_{L+1}(r_L, e; \theta_{L+1})\\
    g(x;\theta) &=& h(e'; \phi),
\end{eqnarray}
to provide the classifier with increased task sensitivity. For network expansion based methods, $e^1, e^2, \dots, e^{t-1}$ will be fixed and only $e^t$ is trained at current task $t$.

\subsection{Training objectives}
Given an input image $x$ with label $y$. Let $g(x;\theta)$ be the current model and $g'(x;\theta')$ the fixed model of the last task, $Y=\sum_{i=1}^t|\mathcal{Y}_i|$, and $Y'=\sum_{i=1}^{t-1}|\mathcal{Y}_i|$ the total number of classes in current task and last task. 

The classification loss is given by:
\begin{align}
    \mathcal{L}_{ce} &= \text{CE}(g(x;\theta), y)
\end{align}
where CE is the cross-entropy function.

In parallel with the classifier $h(x;\theta)$, a $|\mathcal{Y}_t+1|$ way auxiliary classifier $\hat{h}(x;\theta)$ is added over the final feature (in DNE the $e'$) to generate the auxiliary logits:
\begin{align}
    \hat{g}(x;\theta) &= \hat{h}(e'; \hat{\phi})\in\mathbb{R}^{|\mathcal{Y}_t|+1}
\end{align}
where $\hat{\phi}$ is the parameters of the auxiliary classifier.

To supervise this logit, an auxiliary label for input $x$ is given by:
\begin{align}
    \hat{y} &= \left\{\begin{array}{ll}
        1 & \text{if}\quad y \leq Y' \\
        y+1 & \text{else}
    \end{array}
    \right.
\end{align}

The auxiliary labels consider all previous classes as an ``outlier'' class so that the newly added blocks can learn the knowledge that is complementary to previous tasks. This is implemented by the task expert loss:
\begin{align}
    \mathcal{L}_{te} &= \text{CE}(\hat{g}(x;\theta), \hat{y})
\end{align}

The distillation loss is defined as:
\begin{align}
    \mathcal{L}_{dis} &= \text{KL}(\text{SoftMax}(g(x;\theta)[:Y']), \text{SoftMax}(g'(x;\theta)))
\end{align}
where KL is the KL divergence, SoftMax is the SoftMax function and $g(x;\theta)[:Y']$ is the first $Y'$ outputs of $g(x;\theta)$.

The final loss is the weighted average of the previous three losses:
\begin{align}
    \mathcal{L} &= \lambda_1\mathcal{L}_{ce} + \lambda_2\mathcal{L}_{te} + \lambda_3\mathcal{L}_{dis}
\end{align}
where $\lambda_1, \lambda_2$ and $\lambda_3$ are the weights.

\section{Implementation Details}
\subsection{Class balanced tuning}
Similar to \cite{decouple}, we first train the entire network $g(x;\theta)$ with the dataset $D_t$ from current task and the memory buffer$\mathcal{M}$. However, since the size of $D_t$ is typically much larger than $\mathcal{M}$, $g(x;\theta)$ trained on $D_t\cup\mathcal{M}$ is heavily biased towards the classes in current task $t$. To reduce this bias, we subsample the data in $D_t$ to build a subset $D'_t$, such that all the categories have the same number of input data, i.e. $|\{y|(x,y)\in D'_t, y=i\}|=|\{y|(x,y)\in \mathcal{M}, y=j\}|, \forall i,j$. After that, the backbone is fixed, only the last layer $f_{L+1}(r_L, e;\theta_{L+1})$ and the classifier $h(e';\phi)$ is tuned with $D'_t\cup\mathcal{M}$.

\subsection{Dimension of variants of comparison methods}
In Figure~6(a)(b) of the main paper, we compare the accuracy-scale trade-off between DNE and other comparison methods, specifically, we change the feature dimension or backbone network of comparison methods to change their model sizes. The detailed setups are summarized in Table~\ref{tab:setup}, note that, the methods use the identical setups on ImageNet100 and CIFAR100 so we only list the setup of backbone, $N_s$, feature dimension (denoted as Dim in the Table) and the FLOPs $F$ in this setup.

\subsection{Hyperparameters}
We train DNE for 500 epochs and do class balanced tuning for 20 epochs. The learning rate is $2.5\times10^{-4}$ and weight decay is $1\times10^{-6}$, we use SGD optimizer to train our model. The batch is 256 and the model is trained on 4 GPUs in parallel. The memory buffer is set as 2000 and we use the herd selection algorithm\cite{icarl} to update the memory buffer. In the 6-layer transformer, we set the patch size as 4 on CIFAR100 and on ImageNet100 the patch size is 16. The dimension of each head is 32. We empirically set $\lambda_1=\lambda_3=1$ and $\lambda_2=0.1$.

\renewcommand{\thetable}{A}
\begin{table}[]
    \centering
    \begin{tabular}{c|cc|cc}
    \toprule
        Method & Backbone & Dim & $F$\\
        \midrule
        iCaRL\cite{icarl}   & ResNet18      & 768   & 2.50G\\
                & ResNet18       & 864   & 3.16G\\
                & ResNet18       & 1024  & 4.44G\\
    \midrule
        PODNet\cite{podnet}  & ResNet18      & 768   & 2.50G\\
                & ResNet18       & 864   & 3.16G\\
                & ResNet18       & 1024  & 4.44G\\
    \midrule
        Dytox\cite{dytox}   & Transformer       & 512   & 2.46G\\
                & Transformer       & 576   & 3.12G\\
                & Transformer       & 672   & 4.24G\\
    \midrule
        DER\cite{der}     & ResNet18       & 320   & 2.62G\\
                & ResNet18       & 352   & 3.16G\\
                & ResNet18       & 400   & 4.08G\\
    \midrule
        FOSTER\cite{foster}  & ResNet18      & 768   & 2.50G\\
                & ResNet18      & 864   & 3.16G\\
                & ResNet18      & 1024  & 4.44G\\
                & ResNet34      & 512   & 2.32G\\
                & ResNet50      & 512   & 2.62G\\
                & ResNet101     & 512   & 5.04G\\
                & ResNet152     & 512   & 7.48G\\
    \end{tabular}
    \caption{Detailed setups of the comparison methods}
    \label{tab:setup}
\end{table}

\section{Experiments}
\renewcommand{\thefigure}{B}
\begin{figure*}[t]
\centering
\includegraphics[width=\linewidth]{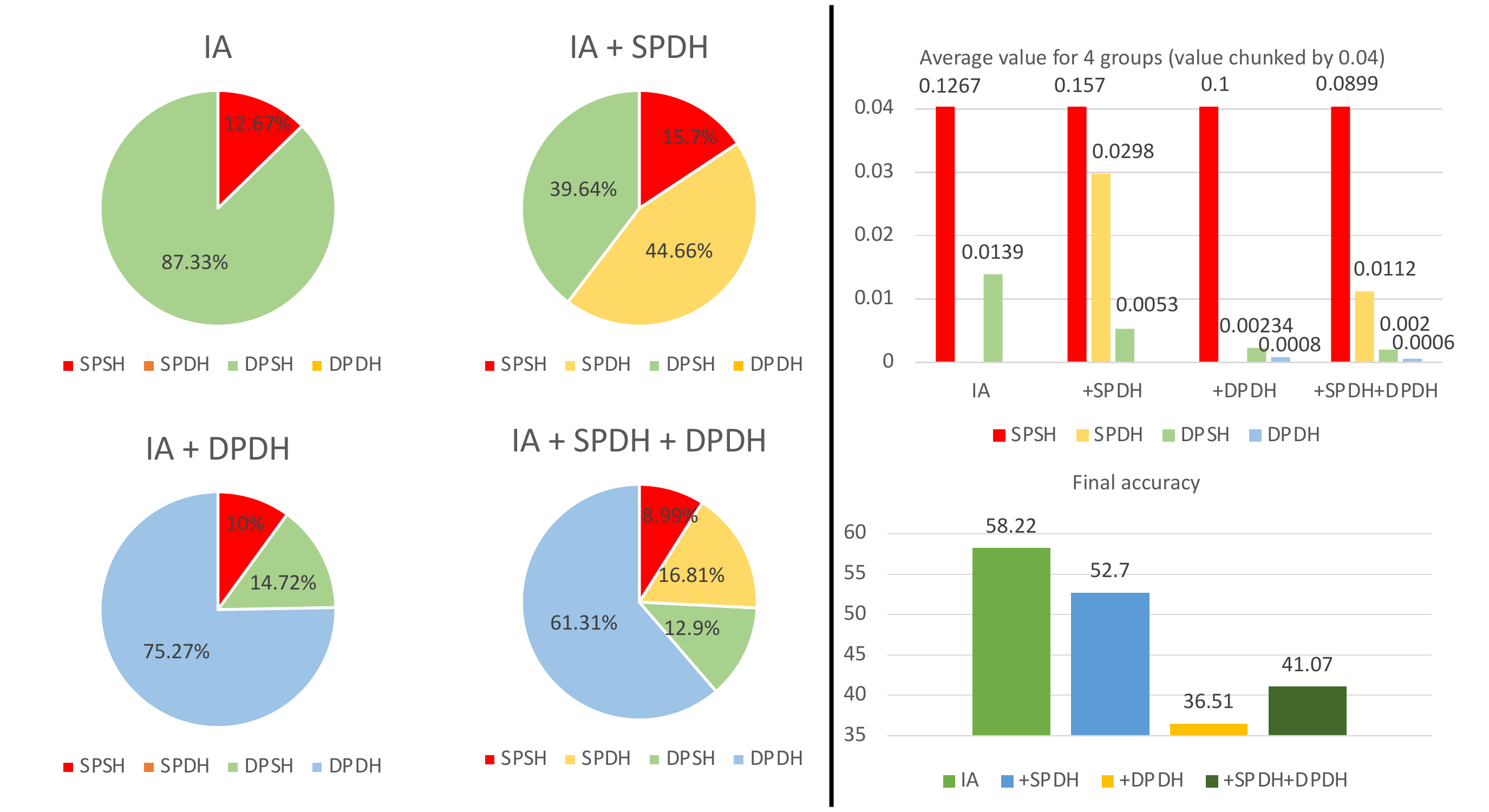} 
\caption{Left: Portion of 4 groups attention within the entire attention matrix for IA model and three variants of STA model. Top-right: Average value of 4 groups attention for IA and three variants of STA. Bottom-right: Final accuracy of IA and three variants of STA.}
\label{fig:analyze}
\end{figure*}

\textbf{Spatial-Task Attention.} We provide a detailed analysis on Spatial-Task Attention(STA) in this experiments. As is illustrated in Figure 2 and 3 of main paper. Attentions in STA are divided into 4 different groups: \textit{Same Patch Same Head} (SPSH), \textit{Different Patch Same Head} (DPSH), \textit{Same Patch Different Head} (SPDH) and \textit{Different Patch Different Head} (DPDH). Standard attention module learns the \textit{Independent Attention} (IA) which includes SPSH and DPSH. This leverages the spatial attention across patches. To implement the joint Spatial-Task Attention, SPDH or DPDH need to be included. This leads to three variants of STA: (IA)+SPDH, (IA)+DPDH and (IA)+SPDH+DPDH. To evalute the effects of these 4 groups of attention in IA and STA, we consider several evaluation metrics. Specifically, the \textit{Portion} of each group of attention, which is the sum of each type of attention divided by the sum of the entire attention matrix. The \textit{Average Value} of each group of attention and the \textit{Final Accuracy} of IA and three variants of STA. The experiments are conducted on CIFAR100 with $N_s=10$.

Figure~\ref{fig:analyze} summarizes these results. In the independent attention, DPSH dominates the attention matrix, so that the spatial attention across different patches is properly learned. However, as SPDH is introduced (IA+SPDH model), the portion of DPSH decreases from 87.33\% to 39.64\%. This is reasonable, as SPDH leverages attention of \textit{exactly the same patch} in different heads, which have very similar representations. As a result, the average value of SPDH (0.0298) is significantly larger than DPSH (0.0053). The attention module is attracted by the cross-head, or cross-task attention so the spatial connections are not well learned. The accuracy also decreases from 58.22\% to 52.70\%. When the DPDH is added (IA+DPDH model and IA+SPDH+DPDH model), the attention matrix is completed dominated by DPDH (75.27\% and 61.31\%). Representations of different patches and different heads are weakly related, this is ture, as the average value of DPDH is merely 0.0008 or 0.0006, significantly smaller than SPDH and DPSH. However, DPDH has much more entries within the attention matrix. Suppose $P$ patches of $H$ heads are considered in STA. SPSH will include $HP$ entry, SPDH includes $HP(H-1)$ entries, DPSH includes $HP(P-1)$ entries and the rest $HP(HP-H-P+1)$ entries all belong to DPDH! In a common setup where $H=16$ and $P=64$, 92\% the entries belong to DPDH. Both SPDH and DPSH are diluted by small-valued but innumerous-numbered DPDH entries. As a result, the final accuracy dropped significantly. Compare to the independent attention model which reaches 58.22\% accuracy, IA+DPDH reaches only 386.51\% and IA+SPDH+DPDH reaches 41.07\%. Based on these observations, we disentangle the spatial and task attention in our proposed DNE model.

\renewcommand{\thefigure}{C}
\begin{figure*}[t]
\centering
\includegraphics[width=0.45\linewidth]{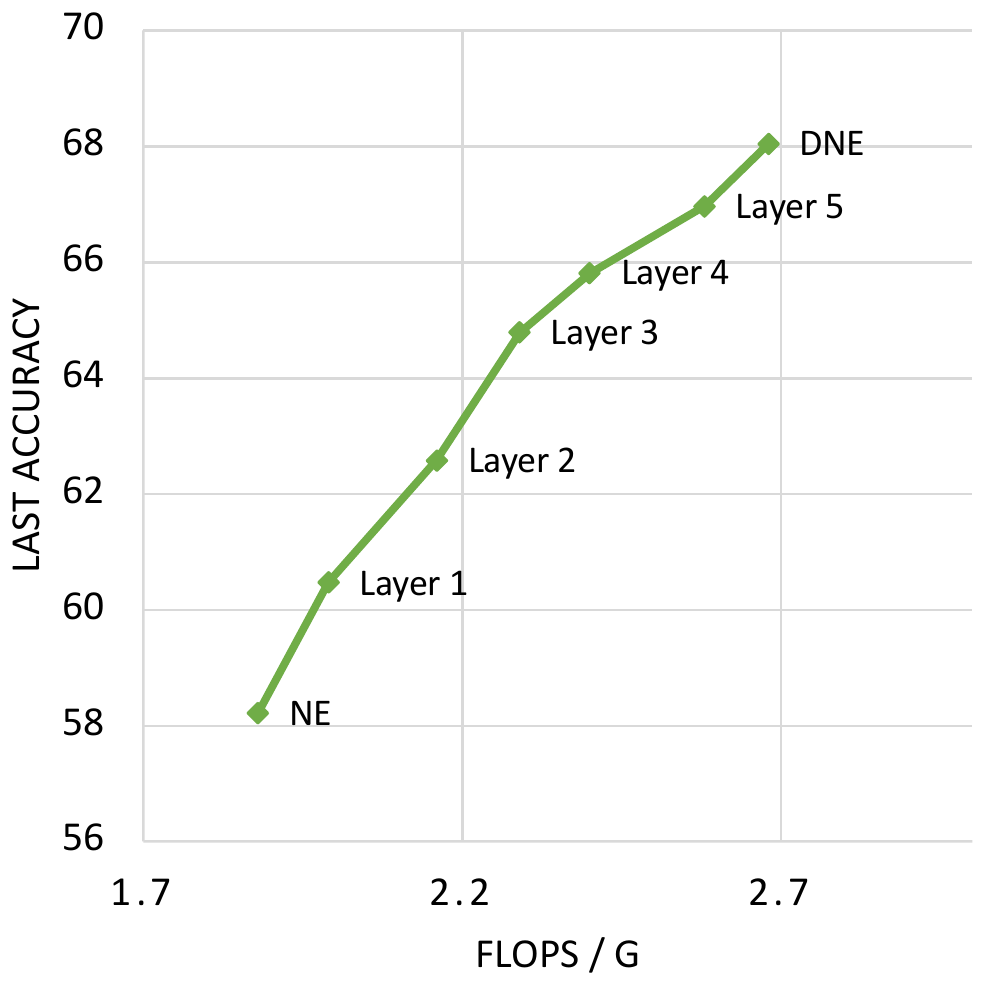}
\includegraphics[width=0.45\linewidth]{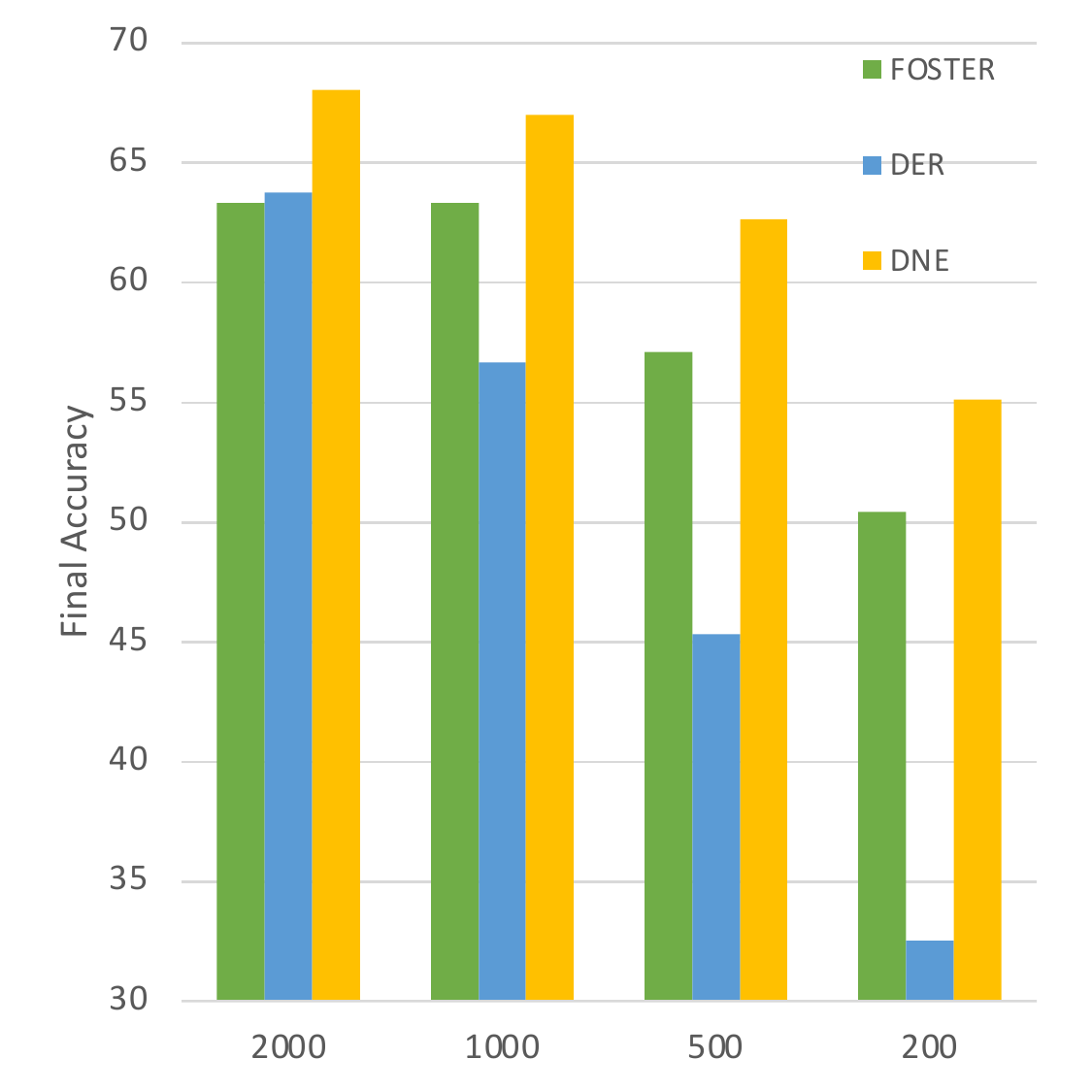} 
\caption{Ablation Study, experiments conducted on CIFAR100 with $N_s=10$. Left: Gradually add CTA across layers. Right: Effect of different memory buffer size.}
\label{fig:ablation}
\end{figure*}

\textbf{CTA vs depth.}
We ablated the impact of adding CTA to each layer of the DNE model. Starting from the NE model (no CTA connections), we gradually add CTA to each layer, from shallow to deep.  Figure~\ref{fig:ablation} left shows that CTA is beneficial at all layers,  with larger gains for shallower layers. While adding CTA to layer 1-3 increases LA by about 2\%, for deep layers the gain is about 1\%. This is not surprising, since the low level features of the shallow layers are more reusable by all tasks. Higher layers have more semantic features, specialized to the task classes. These results show that CTA enables old experts to share knowledge with new experts, and this is useful at all semantic levels.

\textbf{Memory buffer size.} We measure the performances of DNE under different memory settings. The memory buffer size is reduced from 2000 to 200. It can be shown from the right of Figure~\ref{fig:ablation} that DNE consistently outperforms other baseline methods in various memory settings.

\renewcommand{\thetable}{B}
\begin{table}[t]
    \centering
    \begin{tabular}{c|ccc|cc}
        \toprule
        Model & $W_q$ & $W_k$ & $W_v$ & $LA\uparrow$ & $AA\uparrow$\\
        \midrule
            & - & - & s & 63.80 & 70.67\\
        MLP & - & - & f & 67.08 & 73.22\\
        \midrule
        & s & s & s & 64.44 & 71.17\\
        & f & f & s & 64.92 & 71.38\\
        & f & f & f & 67.23 & 73.13\\
        DNE & s & s & f  & {\bf 68.04} & {\bf 73.68}\\
        \bottomrule
    \end{tabular}
    \captionof{table}{Performance on CIFAR100, $N_s=10$, $k=1$, for different configurations of TAB matrices. 's': shared across tasks, 'f': flexible.}
    \label{tab:TABlation}
\end{table}

\textbf{TAB matrices.} The TAB of Figure 5 in main paper has three learned matrices, $W_q, W_k,$ and $W_v$. DNE shares $W_q, W_k,$ across tasks and has flexible $W_v$ per task. This reduces to an MLP across tasks when $W_q, W_k,$ are eliminated and $A_p^{ij}=1, \forall i,j$ in Equation (23) of main paper. We ablated the impact of sharing the different matrices. Table~\ref{tab:TABlation} shows that the DNE configuration has the best trade-off between accuracy and model size.

\renewcommand{\thetable}{C}
\begin{table}[t]
    \centering
    \begin{tabular}{cc|cc}
        \toprule
        \multicolumn{2}{c|}{Cross-task attention in:} & \multirow{2}{*}{$LA\uparrow$}  & \multirow{2}{*}{$AA\uparrow$}\\
        \cline{1-2}
        $l_q,l_k,l_v$ of MHSA & MLP &&\\
        \midrule
        \XSolidBrush    & \XSolidBrush  & 58.22 & 67.63\\
        \Checkmark      & \XSolidBrush  & 66.29 & 72.42\\
        \XSolidBrush    & \Checkmark    & \textbf{68.04} & \textbf{73.68}\\
        \Checkmark      & \Checkmark    & 67.93 & 73.46\\
        \bottomrule
    \end{tabular}
    \caption{Cross-task attention in MHSA and MLP}
    \label{tab:ablation}
\end{table}
\renewcommand{\thetable}{D}
\begin{table}[t]
    \centering
    \begin{tabular}{cc|cc}
        \toprule
        \multicolumn{2}{c|}{Cross-task attention in:} & \multirow{2}{*}{$LA\uparrow$}  & \multirow{2}{*}{$AA\uparrow$}\\
        \cline{1-2}
        $\text{FC}_1$ of MLP & $\text{FC}_2$ of MLP &&\\
        \midrule
        \XSolidBrush    & \XSolidBrush  & 58.22 & 67.63\\
        \Checkmark      & \XSolidBrush  & 66.82 & 73.07\\
        \XSolidBrush    & \Checkmark    & 66.74 & 72.77\\
        \Checkmark      & \Checkmark    & \textbf{68.04} & \textbf{73.68}\\
        \bottomrule
    \end{tabular}
    \caption{Cross-task attention in different linear layers MLP, $\text{FC}_1$ is the first linear layer of MLP and $\text{FC}_2$ is the secound linear layer of MLP}
    \label{tab:mlp_layers}
\end{table}

\textbf{Cross-task attention in MHSA.} MHSA and MLP are the main components of the transformer block. MHSA mainly learns spatial attentions between different patches while MLP fuse the features of different channels. In DNE, cross-task attention(CTA) is only equipped in MLP. However, there are still linear layers in the MHSA block, specifically the three linear layers $l_q, l_k, l_v$ to generate query, key and value vectors in the self-attention blocks. By applying CTA to these linear layers, the MHSA could be able to first learn a feature that encodes information across tasks and then learn the spatial relationships. In Table~\ref{tab:ablation}, we evaluate the effects of CTA in the linear layers of MHSA and MLP. The experiments are conducted on CIFAR100 with $N_s=10$.

Using CTA in MHSA and MLP can both significantly improve the performances. CTA in MHSA(MLP) increases the last accuracy by 8.07\%(9.82\%) and the average incremental accuracy by 4.79\%(6.05\%), compared to the model without CTA. But using CTA in both MHSA and MLP does not further boost the performances. Linear layers in MHSA and the MLP are doing similar things while MHSA is more focusing on spatial connections. Thus it is more natural to implement the CTA in MLP block only.

\textbf{Cross-task attention in different layers of MLP.} MLP has two linear layers in which CTA can be added. In Table~\ref{tab:mlp_layers} we evaluate the effects of CTA in these two linear layers. Experiments are conducted on CIFAR100 with $N_s=10$. The results show that using CTA in either the first or the second linear layer can improve the last accuracy by about 8.5\% and the average incremental accuracy by about 5.5\%. But by using CTA in both linear layers, the performances can be further improved. In DNE we use CTA in both linear layers to reach better performances. 

\end{document}